\documentclass[a4paper,conference]{IEEEtran}
\usepackage{cite}
\usepackage{amsmath}
\usepackage{algorithmic}
\usepackage{epsfig} 
\usepackage{array}
\usepackage{graphicx}
\usepackage{amsmath}
\graphicspath{ {./images/} }
\usepackage{stfloats}
\usepackage[most]{tcolorbox}
\usepackage{url}
\providecommand{\keywords}[1]
{
 \small	
 \textbf{\textit{Keywords---}} #1
}
\usepackage{hhline}

\begin{document}

\title{CDeC-Net: Composite Deformable Cascade Network for Table Detection in Document Images}

\author{\IEEEauthorblockN{Madhav Agarwal}
\IEEEauthorblockA{CVIT, IIIT, Hyderabad, India\\
madhav14130@gmail.com}
\and
\IEEEauthorblockN{Ajoy Mondal}
\IEEEauthorblockA{CVIT, IIIT, Hyderabad, India \\
ajoy.mondal@iiit.ac.in}
\and
\IEEEauthorblockN{C. V. Jawahar}
\IEEEauthorblockA{CVIT, IIIT, Hyderabad, India \\
jawahar@iiit.ac.in}}
\maketitle

\begin{abstract}
Localizing page elements/objects such as tables, figures, equations, etc. is the primary step in extracting information from document images. We propose a novel end-to-end trainable deep network, ({\sc \textbf{cd}}e{\sc \textbf{c-n}}et) for detecting tables present in the documents. The proposed network consists of a multistage extension of Mask {\sc \textbf{r-cnn}} with a dual backbone having deformable convolution for detecting tables varying in scale with high detection accuracy at higher {\sc \textbf{i}}o{\sc \textbf{u}} threshold. We empirically evaluate {\sc \textbf{cd}}e{\sc \textbf{c-n}}et on all the publicly available benchmark datasets --- {\sc \textbf{icdar-2013}}, {\sc \textbf{icdar-2017}}, {\sc \textbf{icdar-2019}}, {\sc \textbf{unlv}}, {\sc \textbf{m}}armot, {\sc \textbf{p}}ub{\sc \textbf{l}}ay{\sc \textbf{n}}et, and {\sc \textbf{t}}able{\sc \textbf{b}}ank --- with extensive experiments.

Our solution has three important properties: (i) a single trained model {\sc \textbf{cd}}e{\sc \textbf{c-n}}et$^{\ddagger}$ performs well across all the popular benchmark datasets; (ii) we report excellent performances across multiple, including higher, thresholds of {\sc \textbf{i}}o{\sc \textbf{u}}; (iii) by following the same protocol of the recent papers for each of the benchmarks, we consistently demonstrate the superior quantitative performance. Our code and models will be publicly released for enabling the reproducibility of the results.

\end{abstract}

\keywords{Page object, table detection, Cascade Mask {\sc r-cnn}, deformable convolution, single model.}

\ifCLASSOPTIONpeerreview
 \begin{center} \bfseries EDICS Category: 3-BBND \end{center}
\fi
\IEEEpeerreviewmaketitle

\section{Introduction}

Rapid growth in information technology has led to an exponential increase in the production and storage of digital documents over the last few decades. Extracting information from such a large corpus is impractical for human. Hence, useful information could be lost or not utilized over time. Digital documents have many other page objects (such as tables and figures) beyond text. These page objects also show wide variations in
their appearance. Therefore any attempt to detect page objects such as tables need to be generic and applicable across wide variety of documents and use cases. In this paper, we are interested in detection of tables. 
It is well known~\cite{gilani2017table,schreiber2017deepdesrt,siddiqui2018decnt,sun2019faster,vo2018ensemble,arif2018table,li2019tablebank,younas2019ffd,zhong2019publaynet,saha2019graphical,casado2019benefits} that the localisation of tables and other page element is challenging due to the high degree of intra-class variability (due to different layouts of the table, inconsistent use of ruling lines). The presence of inter-class similarity (graphs, flowcharts, figures having a large number of horizontal and vertical lines which resembles to table) adds further challenges. 

Table detection is still a challenging problem in the research community. This is an active area of research~\cite{siddiqui2018decnt,sun2019faster,vo2018ensemble,arif2018table,li2019tablebank,younas2019ffd,zhong2019publaynet,saha2019graphical,casado2019benefits}. However, we observe that most of these attempts develop different table detection solutions for different datasets. We argue that this may be the time to consider the possibility of a single solution (say a trained model) that works across wide variety of documents. We provide a single model {\sc cd}e{\sc c-n}et$^{\ddagger}$ trained with {\sc iiit-ar-13k} dataset~\cite{ajoy2020_das} and evaluate on popular benchmark datasets. Table~\ref{single_comparion_table} shows the comparison with the state-of-the-art techniques for respective datasets. We observe from the table that our single model {\sc cd}e{\sc c-n}et$^{\ddagger}$ performs better than state-of-the-art techniques for {\sc icdar-2019} (c{\sc td}a{\sc r})~\cite{gao2019icdar}, {\sc unlv}~\cite{shahab2010open}, and {\sc p}ub{\sc l}ay{\sc n}et~\cite{zhong2019publaynet} dataset. In case of {\sc icdar-2013}~\cite{gobel2013icdar}, {\sc icdar-pod-2017}~\cite{gao2017icdar2017}, {\sc m}armot~\cite{fang2012dataset}, and {\sc t}able{\sc b}ank~\cite{li2019tablebank}, single model {\sc cd}e{\sc c-n}et$^{\ddagger}$ obtains comparable results to the state-of-the-art techniques. By following the same protocol of the state of the art papers, we also report superior performance consistently across all the datasets, as presented in Table~\ref{comparion_table} and discussed later in this paper.

\begin{table}[ht!]
\addtolength{\tabcolsep}{-2.0pt}
\begin{center}
\begin{tabular}{|l|l|c c c c|} \hline
\textbf{Dataset} &\textbf{Method} &\multicolumn{4}{|c|}{\textbf{Score}} \\ \cline{3-6}
 &  &\textbf{R}$\uparrow$ &\textbf{P}$\uparrow$ &\textbf{F1}$\uparrow$ &\textbf{mAP}$\uparrow$ \\ \hline 
{\sc icdar-2013} &{\sc d}e{\sc cnt}~\cite{siddiqui2018decnt} &\textbf{0.996}$^{*}$ &\textbf{0.996}$^{*}$ &\textbf{0.996}$^{*}$ &- \\
 &{\sc cd}e{\sc c-n}et$^{\ddagger}$ (our) &0.942 &0.993 &0.968 &\textbf{0.942} \\ \hline
{\sc icadr-2017} &{\sc yolo}v3~\cite{huang2019yolo} &\textbf{0.968} &\textbf{0.975} &\textbf{0.971} &- \\
&{\sc cd}e{\sc c-n}et$^{\ddagger}$ (our) &0.899	&0.969 &0.934 &0.880 \\  \hline
{\sc icadr-2019} &{\sc t}able{\sc r}adar~\cite{gao2019icdar} &\textbf{0.940} &0.950 &0.945 &- \\
 &{\sc cd}e{\sc c-n}et$^{\ddagger}$ (our) &0.930 &\textbf{0.971} &\textbf{0.950} &\textbf{0.913} \\ \hline 
 {\sc unlv} &{\sc god}~\cite{saha2019graphical} &0.910 &0.946 &0.928 &- \\ 
  &{\sc cd}e{\sc c-n}et$^{\ddagger}$ (our) &\textbf{0.915} &\textbf{0.970} &\textbf{0.943} &\textbf{0.912} \\ \hline
{\sc m}armot &{\sc d}e{\sc cnt}~\cite{siddiqui2018decnt} &\textbf{0.946} &0.849 &\textbf{0.895} &- \\ 
&{\sc cd}e{\sc c-n}et$^{\ddagger}$ (our) &0.779 &\textbf{0.943} &0.861 &\textbf{0.756} \\ \hline
{\sc t}able{\sc b}ank &Li et al.~\cite{li2019tablebank} &\textbf{0.975} &0.987 &\textbf{0.981} &- \\
 &{\sc cd}e{\sc c-n}et$^{\ddagger}$ (our) &0.970 &\textbf{0.990} &0.980 &\textbf{0.965} \\ \hline
{\sc p}ub{\sc l}ay{\sc n}et &{\sc m-rcnn}~\cite{zhong2019publaynet} &- & -&- &0.960 \\ 
 &{\sc cd}e{\sc c-n}et$^{\ddagger}$ (our) & \textbf{0.975} & \textbf{0.993} & \textbf{0.984} & \textbf{0.978}  \\ \hline
 \end{tabular}
\end{center}
\caption{Illustrates comparison between our single model {\sc cd}e{\sc c-n}et$^{\ddagger}$ and state-of-the-art techniques on existing benchmark datasets. We create the single model {\sc cd}e{\sc c-n}et$^{\ddagger}$ by training {\sc cd}e{\sc c-n}et with {\sc iiit-ar-13k} and fine-tuning with training set of respective datasets. \textbf{*:} indicates the authors reported 0.996 in table however in discussion they mentioned 0.994. \label{single_comparion_table}}
\end{table}
 
Early attempts in localizing tables are based on meta-data extraction and exploitation of the semantic information present in the tables~\cite{Itonori1993TableSR,Kieninger1998TableSR,Tupaj96extractingtabular}. However, the absence of meta-data in  the case of scanned documents makes these methods futile. In recent years, researchers employ deep neural networks~\cite{gilani2017table,schreiber2017deepdesrt,siddiqui2018decnt,sun2019faster,vo2018ensemble,arif2018table,li2019tablebank,younas2019ffd,zhong2019publaynet,saha2019graphical,casado2019benefits} in an attempt to provide a  generic solution for localizing page objects, specifically tables from document images. Siddiqui et al.~\cite{siddiqui2018decnt} provide state-of-the-art performance on  many benchmark datasets by incorporating deformable convolutions~\cite{dai2017deformable} in their network. However, even their work fails to provide a single model that achieves state-of-the-art performance on all the existing benchmark datasets. In general, the existing deep learning models are trained on a single {\sc i}o{\sc u} threshold, commonly 0.5, following the practice followed in computer vision literature.  This leads to a noisy table detection at higher threshold value during evaluation. This is a drawback of the existing table detection techniques. Liu et al. discuss in~\cite{liu2019cbnet} that generally, a {\sc cnn} based object detector uses a backbone network to extract features for detecting objects. These backbones are usually designed for the image classification task and  are pre-trained on either {\sc i}mage{\sc n}et~\cite{deng2009imagenet} or {\sc ms-coco}~\cite{lin2014microsoft} datasets. Hence, directly employing them to extract features for table detection~\cite{gilani2017table,schreiber2017deepdesrt,siddiqui2018decnt,sun2019faster,vo2018ensemble,arif2018table,li2019tablebank,younas2019ffd,zhong2019publaynet,saha2019graphical,casado2019benefits} may result in sub-optimal performance. Training a more powerful backbone is also expensive. This is a major bottleneck of these existing table detection techniques.         
 
To address the issues mentioned above, we propose a composite deformable cascade network, called as {\sc cd}e{\sc c-n}et, to detect tables more accurately present in document images. The proposed {\sc cd}e{\sc c-n}et consists of a multi-stage object detection architecture, cascade Mask {\sc r-cnn}~\cite{cai2019cascade}. The cascade Mask {\sc r-cnn} network is composed of a sequence of detectors trained with increasing {\sc i}o{\sc u} thresholds to address the problem of noisy detection at higher threshold. Inspired by~\cite{liu2019cbnet} we use composite backbone, which consists of multiple identical backbones having composite connections between neighbor backbones, in {\sc cd}e{\sc c-n}et to improve detection accuracy. We also incorporate deformable convolution~\cite{dai2017deformable} in the backbones to model geometric transformations. We extensively evaluate {\sc cd}e{\sc c-n}et on publicly available benchmark datasets --- {\sc icdar-2013}, {\sc icdar-pod-2017}, {\sc unlv}, Marmot, {\sc icdar-2019} (c{\sc td}a{\sc r}), {\sc t}able{\sc b}ank, and {\sc p}ub{\sc l}ay{\sc n}et under various existing experimental environments. The extensive experiments show that {\sc cd}e{\sc c-n}et achieves state-of-the-art performance on all existing benchmark datasets except {\sc icdar-2017}. We also achieve high accuracy and more tight bounding box detection at higher {\sc i}o{\sc u} threshold than the previous benchmark results. 

We summarise our main contributions as follows: 

\begin{itemize}

\item We presents an end-to-end trainable deep architecture, {\sc cd}e{\sc c-n}et which consists of {\sc c}ascade {\sc m}ask {\sc r-cnn} containing composite backbones with deformable convolution to detect tables more accurately in document images. 

\item We provide a single model trained on {\sc iiit-ar-13k} and achieve very close competitive results to the state-of-the-art techniques on all existing benchmark datasets (Refer Table~\ref{single_comparion_table}).

\item We achieve state-of-the-art results on publicly available benchmark datasets except {\sc icdar-2017} (Refer Table~\ref{comparion_table}).

\end{itemize}

\section{Related Work} \label{Related Work}

Table detection is an essential step towards document analysis. Over the times, many researchers have contributed to the detection of tables in documents of varying layouts. Initially, the researchers have proposed several approaches based on heuristics or meta-data information to solve this particular problem~\cite{Itonori1993TableSR,chandran1993structural,hirayama1995method,green1995recognition,Tupaj96extractingtabular,hu1999medium,gatos2005automatic,shafait2010table}. Later, the researchers explore machine learning, more specifically deep learning, to make the solution generic~\cite{gilani2017table,schreiber2017deepdesrt,siddiqui2018decnt,sun2019faster,vo2018ensemble,arif2018table,li2019tablebank,younas2019ffd,zhong2019publaynet,saha2019graphical,casado2019benefits}. 

\subsection{Rule Based Approaches}

The research on table detection in document images was started in 1993. In the beginning, Itonori~\cite{Itonori1993TableSR} proposed a rule-based approach that led to the text-block arrangement and ruled line position to localize the table in the documents. At the same time, Chandran and Kasturi~\cite{chandran1993structural} developed a table detection approach based on vertical and horizontal lines. Following these, several research works~\cite{hirayama1995method,green1995recognition,Tupaj96extractingtabular,hu1999medium,gatos2005automatic,shafait2010table} have been done for table detection using improved heuristic rules. Though these methods perform well on the documents having limited variation in layouts, they need more manual efforts to find a better heuristic rule. Moreover, rule-based approaches fail to obtain generic solutions. Therefore, it is necessary to employ machine learning approaches to solve the table detection problem.  

\subsection{Learning Based Approaches}

Statistical learning approaches have been proposed to alleviate the problems mentioned earlier in table detection. Kieninger and Dengel~\cite{kieninger1998t} applied an unsupervised learning approach for the table detection task. This method significantly differs from the previous rule-based approaches~\cite{hirayama1995method,green1995recognition,Tupaj96extractingtabular,hu1999medium,gatos2005automatic,shafait2010table} as it uses a clustering of given word segments. Cesarini et al.~\cite{cesarini2002trainable} used a supervised learning approach using a hierarchical representation based on the {\sc mxy} tree. This particular method detects the table with different features by maximizing the performance on a particular training set. Later, the solution of the table detection problem is formulated using various machine learning problems such as (i) sequence labeling~\cite{e2009learning}, (ii) {\sc svm} with various hand-crafted features~\cite{kasar2013learning}, and (iii) ensemble of various models~\cite{fan2015table}. Learning methods improve table detection accuracy significantly.

\begin{table}[h!]
\addtolength{\tabcolsep}{-4.0pt}
\begin{center}
\begin{tabular}{|l|l|r|r|r|} \hline
\textbf{Dataset} &\textbf{Category Label} &\textbf{Training} &\textbf{Validation} &\textbf{Test}  \\ 
                 &   &\textbf{Set} &\textbf{Set} &\textbf{Set} \\ \hline
{\sc icdar-2013}     &1: T &170 &  &238 \\
{\sc icdar-pod-2017} &3: T, F, and E &1600 & &817 \\
{\sc unlv}           &1: T &   &   &424 \\
Marmot               &1: T &2K &  &  \\
{\sc icdar-2019} (c{\sc td}a{\sc r}) &1: T &1200 & &439 \\
{\sc t}able{\sc b}ank-{w}ord\footnotemark[1]           &1: T &163K &1K &1k \\
{\sc t}able{\sc b}ank-{\sc l}a{\sc t}e{\sc x}\footnotemark[1]  &1: T &253K &1K &1k \\
{\sc t}able{\sc b}ank-both\footnotemark[1]           &1: T &417K &2K &2k \\
{\sc p}ub{\sc l}ay{\sc n}et\footnotemark[1]          &5: T, F, TL, TT, and LT &340K &11K &11K \\
{\sc iiit-ar-13k}                    &5: T, F, NI, L, and S   &9K &2K &2k \\ \hline
\end{tabular}
\end{center}
\caption{Statistics of datasets. \textbf{T:} indicates table. \textbf{F:} indicates figure. \textbf{E:} indicates equation. \textbf{NI:} indicates natural image. \textbf{L:} indicates logo. \textbf{S:} indicates signature. \textbf{TL:} indicates title. \textbf{TT:} indicates text. \textbf{LT:} indicates list. \label{table_statistics_dataset}}
\end{table}
\footnotetext[1]{Ground truth bounding boxes are annotated automatically.}

The success of deep convolutional neural network ({\sc cnn}) in the field of computer vision, motivates researchers to explore {\sc cnn} for localizing tables in the documents. It is a data-driven method and has advantages --- (i) it is robust to document types and layouts, and (ii) it reduces the efforts of hand-crafted feature engineering in {\sc cnn}. Initially, Hao et al.~\cite{hao2016table} used {\sc cnn} to classify tables like structure regions extracted from {\sc pdf}s using heuristic rule into two categories - table and non-table. The major drawbacks of this method are (i) use of the heuristic rule to extract table like region, and (ii) work on only non-raster {\sc pdf} documents. The researchers explore various natural scene object detectors --- Fast {\sc r-cnn}~\cite{girshick2015fast} in~\cite{vo2018ensemble}, Faster {\sc r-cnn}~\cite{ren2015faster} in~\cite{gilani2017table,schreiber2017deepdesrt,siddiqui2018decnt,vo2018ensemble,arif2018table,li2019tablebank,sun2019faster,younas2019ffd,zhong2019publaynet}, Mask {\sc r-cnn}~\cite{he2017mask} in~\cite{younas2019ffd,saha2019graphical,zhong2019publaynet,casado2019benefits}, {\sc yolo}~\cite{redmon2016you} in~\cite{casado2019benefits} to localize page objects more specifically tables in the document images. All these methods are  data-driven and do not require any heuristics or meta-data to extract table like region similar to~\cite{hao2016table}.  

\begin{figure*}[ht!]
\centerline{
\tcbox[sharp corners, size = tight, boxrule=0.2mm, colframe=black, colback=white]{
\psfig{figure=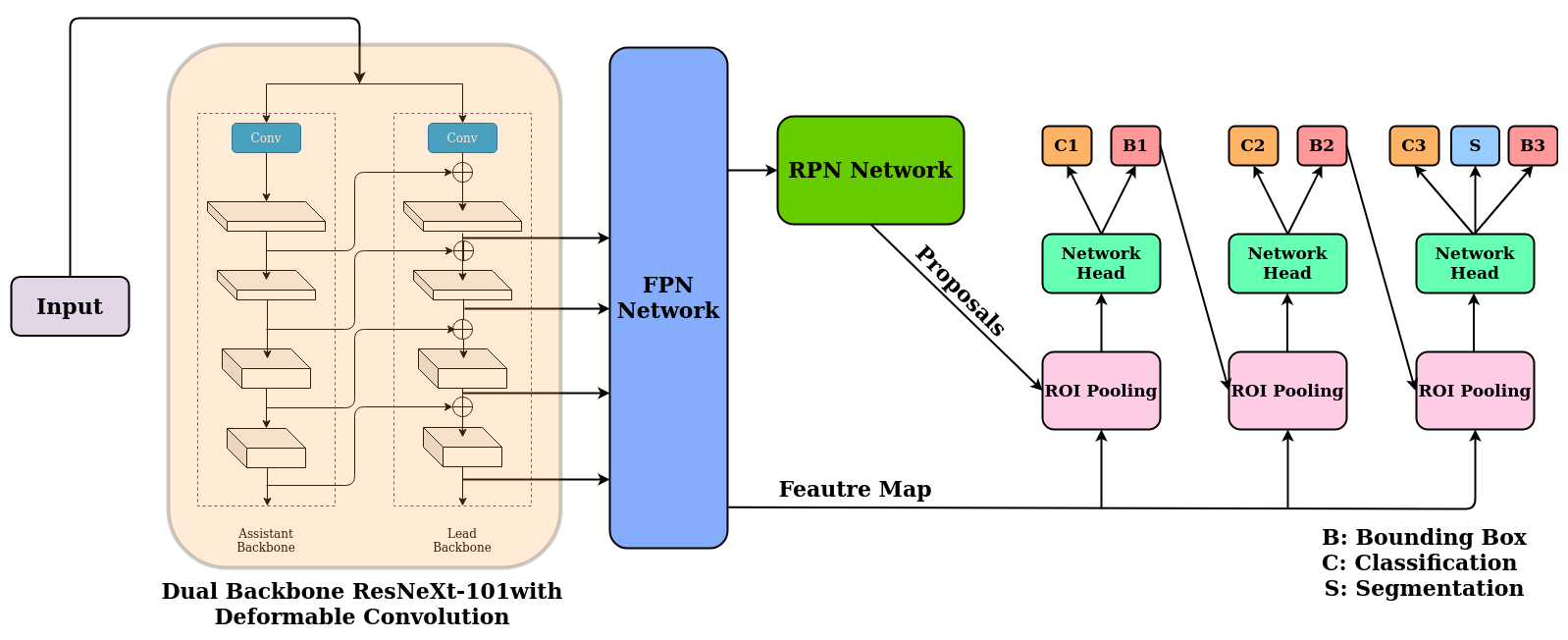, width=0.9\textwidth,height=0.23\textwidth}}}
\caption{Illustration of the proposed {\sc cd}e{\sc c-n}et which is compose of cascade Mask {\sc r-cnn} with composite backbone having deformable convolution instead of conventional convolution. \label{fig:final_architecture}}
\end{figure*}

Gilani et al.~\cite{gilani2017table} used Faster {\sc r-cnn} to detect tables in the document images. Instead of the original document image, distance transformed image is taken as input to easily fine-tune the pre-trained model to work on various types of document images. In the same direction, the transformed document image is taken as input to Faster {\sc r-cnn} model for detecting tables~\cite{arif2018table}; and figures and mathematical equations~\cite{younas2019ffd} present in document images. Saha et al.~\cite{saha2019graphical} experimentally established that Mask {\sc r-cnn} performs better than Faster {\sc r-cnn} for detecting graphical objects in the document images. Zhong et al.~\cite{zhong2019publaynet} also experimentally established that Mask {\sc r-cnn} performs better than Faster {\sc r-cnn} for extracting semantic regions from the documents. The performance of Faster {\sc r-cnn} is reduced when documents contain large scale variate tables. Siddiqui et al.~\cite{siddiqui2018decnt} incorporated deformable {\sc cnn} in Faster {\sc r-cnn} to adapt the different scales and transformations which allows the model to detect scale variate tables accurately. Sun et al.~\cite{sun2019faster} combined the corner information with the detected table region by Faster {\sc r-cnn} to refine the boundaries of the detected tables to reduce false positives. 
It is observed that every detection method is sensitive to a certain type of object. Vo et al.~\cite{vo2018ensemble} combine outputs of two object detectors --- Fast {\sc r-cnn} and Faster {\sc r-cnn} in order to exploit the advantages of the two models for page object detection. Due to the limited number of images in the existing training set, it is challenging to train such a detection model for table detection. Fine-tune is one solution to such a problem. In~\cite{casado2019benefits}, the authors discuss the benefit of fine-tuning from a close domain on four different object detection models --- Mask {\sc r-cnn}~\cite{he2017mask}, {\sc r}etina{\sc n}et~\cite{lin2017focal}, {\sc ssd}~\cite{liu2016ssd} and {\sc yolo}~\cite{redmon2016you}. The experiments highlight that the close domain fine-tuning approach avoids over-fitting, solves the problem of having a small training set, and improves detection accuracy.

\subsection{Related Datasets}

Various benchmark datasets --- {\sc icdar-2013}~\cite{gobel2013icdar}, {\sc icdar-pod-2017}~\cite{gao2017icdar2017}, {\sc unlv}~\cite{shahab2010open}, {\sc m}armot~\cite{fang2012dataset}, {\sc icdar-2019} (c{\sc td}a{\sc r})~\cite{gao2019icdar}, {\sc t}able{\sc b}ank~\cite{li2019tablebank}, {\sc p}ub{\sc l}ay{\sc n}et~\cite{zhong2019publaynet}, and {\sc iiit-ar-13k}~\cite{ajoy2020_das} are publicly available for table detection tasks. Table~\ref{table_statistics_dataset} shows the statistics of these datasets. Among them, {\sc icdar-2013}, {\sc unlv}, {\sc m}armot, {\sc icdar-2019}, {\sc t}able{\sc b}ank are popularly used for table detection while {\sc icdar-pod-2017}, {\sc p}ub{\sc l}ay{\sc n}et, and {\sc iiit-ar-13k} datasets for various page object (including table) detection task. We use all datasets for our experiments.          

\section{CDeC-Net: Composite Deformable Cascade Network}

The success of deep convolution neural networks ({\sc cnn})s for solving various computer vision problems inspire researchers to explore and design models for detecting tables in document images~\cite{gilani2017table,schreiber2017deepdesrt,siddiqui2018decnt,sun2019faster,vo2018ensemble,arif2018table,li2019tablebank,younas2019ffd,zhong2019publaynet,saha2019graphical,casado2019benefits}. All these deep models provide high table detection accuracy. However, all these table detection models suffer from the following shortcomings --- (i) all existing table detection networks use a backbone to extract features for detecting tables, which is usually designed for image classification tasks and pre-trained on {\sc i}mage{\sc n}et dataset. Since almost all of the existing backbone networks are originally designed for the image classification task, directly applying them to extract features for table detection may result in sub-optimal performance. A more powerful backbone is needed to extract more representational features and improve the detection accuracy. However it is very expensive to train a deeper and powerful backbone on {\sc i}mage{\sc n}et and get better performance. (ii) {\sc cnn}s have limitations to model large transformation due to the fixed geometric structures of {\sc cnn} modules --- a convolution filter samples the input feature map correspond to a fixed location, a pooling layer reduces the spatial resolution at a fixed ration and a {\sc r}o{\sc i} into a fixed spatial bin, etc. This leads to the lack of handling the geometric transformations. (iii) All these table detectors use the intersection over union ({\sc i}o{\sc u}) threshold to define positives, negatives, and finally, detection quality. They commonly use a threshold of 0.5, which leads to noisy (low-quality) detection and frequently degrades the performance for higher thresholds. The major hindrance in training a network at higher {\sc i}o{\sc u} threshold is the reduction of positive training samples with increasing {\sc i}o{\sc u} threshold. All these issues are also a bottleneck of {\sc cnn}s based object detection techniques~\cite{girshick2015fast,ren2015faster,he2017mask} in natural scene images.

Over the time, various solutions~\cite{liu2019cbnet,dai2017deformable,cai2019cascade} are proposed to handle the above stated problems for object detection in natural images. Lie et al.~\cite{liu2019cbnet} proposed {\sc cbn}et which comprises of stacking multiple identical backbones by creating composite connections between them. It helps in creating a more powerful backbone for feature extraction without much additional computational cost. Dai et al.~\cite{dai2017deformable} introduced deformable convolution in the object detection network to make it more scale-invariant. It captures the features using a variable receptive field and makes detection independent of the fixed geometric transforms. Cai and Vasconcelos~\cite{cai2019cascade} proposed a multi-stage object detection architecture in which subsequent detectors are trained with increasing {\sc i}o{\sc u} thresholds to solve the last problem. The output of one detector is feed as an input to the subsequent detector, maintaining the number of positive samples at higher thresholds.

Inspired by the solutions provided by~\cite{liu2019cbnet,dai2017deformable,cai2019cascade} for issues discussed earlier in natural scene images, we propose a novel architecture {\sc cd}e{\sc c-n}et for detecting tables accurately in the document images. It is composed of {\sc c}ascade {\sc m}ask {\sc r-cnn} with a composite backbone having deformable convolution filters instead of conventional convolution filters. Figure~\ref{fig:final_architecture} displays an overview of our proposed architecture for table localization in document images. We discuss each component of {\sc cd}e{\sc c-n}et in detail:

\subsection{Cascade Mask R-CNN}

Cai and Vasconcelos~\cite{cai2019cascade} proposed Cascade {\sc r-cnn} which is a multi-stage extension of Faster {\sc r-cnn}~\cite{ren2015faster}. Cascade Mask {\sc r-cnn} has a similar architecture as Cascade {\sc r-cnn}, but along with an additional segmentation branch, denoted by '{\sc s}', for creating masks of the detected objects. {\sc cd}e{\sc c-n}et comprises of a sequence of three detectors trained with increasing {\sc i}o{\sc u} thresholds of 0.5, 0.6, and 0.7, respectively. The proposals generated by {\sc rpn} network are passed through {\sc roi} pooling layer. The network head takes {\sc roi} features as input and makes two predictions --- classification score ({\sc c}) and bounding box regression ({\sc b}). The output of one detector is used as a training set for the next detector. The deeper detector stages are more selective against close false positives. Each regressor is optimized for the bounding box distribution generated by the previous regressor, rather than the initial distribution. The bounding box regressor trained for a certain {\sc i}o{\sc u} threshold, tends to produce bounding boxes of higher {\sc i}o{\sc u} threshold. It helps in re-sampling an example distribution of higher {\sc i}o{\sc u} threshold and uses it to train the next stage. Hence, it results in a uniform distribution of training samples for each stage of detectors and enabling the network to train on higher {\sc i}o{\sc u} threshold values. 

\subsection{Composite Backbone}

We use a dual backbone based architecture~\cite{liu2019cbnet} which creates a composite connection between the parallel stages of two adjacent {\sc r}es{\sc n}e{\sc x}t-101 backbones (one is called assistant backbone and other is called lead backbone). The assistant backbone's high-level output features are fed as an input to the corresponding lead backbone's stage.
In a conventional network, the output (denoted by $x^l$) of previous \emph{l}-1 stages  is feed as input to the \emph{l}-th stage, given by:
\begin{equation}
    x^l=F^l(x^{l} - 1) , l\geq 2.
\end{equation}
where $F^l$(.) is a non-linear transformation operation of \emph{l}-th stage.
However, our network takes input from previous stages as well as parallel stage of assistant backbone. For a given stage \emph{l} of lead backbone(bl), input is a combination of the output of previous \emph{l}-1 stages of lead backbone and parallel \emph{l}-th stage of assistant backbone(ba), given by:
\begin{equation}
    x^{l}_{bl} = F^{l}_{bl}(x^{l-1}_{k}  + g(x^{l}_{ba})), l\geq2,
\end{equation}
where g(.) represents composite connection. It helps the lead backbone to take advantage of the features learned by the assistant backbone. Finally, the output of the lead backbone is used for further processing in the subsequent network. 

\subsection{Deformable Convolution}

The commonly used backbone, {\sc r}es{\sc n}e{\sc x}t architectures, have conventional convolution operation, in which the effective receptive field of all the neurons in a given layer is same. The grid points are generally confined to a fix 3$\times$3 or 5$\times$5 square receptive field. It performs well for layers at the lower hierarchy, but when the objects appear at the arbitrary scales and transformations, generally at the higher-level, the convolution operation does not perform well in capturing the features. We replace the fixed receptive field {\sc cnn} with deformable {\sc cnn}~\cite{dai2017deformable} in each of our dual backbone architectures. The gird is deformable as each grid point can be moved by a learnable offset. In a conventional convolution, we sample over the input feature map x using a regular grid R, given by

\begin{equation}
  y(p_0) = \sum_{p_n\in R} w(p_n).x(p_0 +p_n).  
\end{equation}
Whereas in a deformable convolution, for each location $p_o$ on the output feature map y,  we augment the regular grid using the offset $\Delta p_n$ such that $\{\Delta p_n |n= 1,...,N\}$, where $N=|R|$, given by

\begin{equation}
y(p_0) = \sum_{p_n\in R} w(p_n).x(p_0 +p_n +\Delta p). 
\end{equation}

\begin{figure}[ht!]
\centerline{
\tcbox[sharp corners, size = tight, boxrule=0.2mm, colframe=black, colback=white]{
\psfig{figure=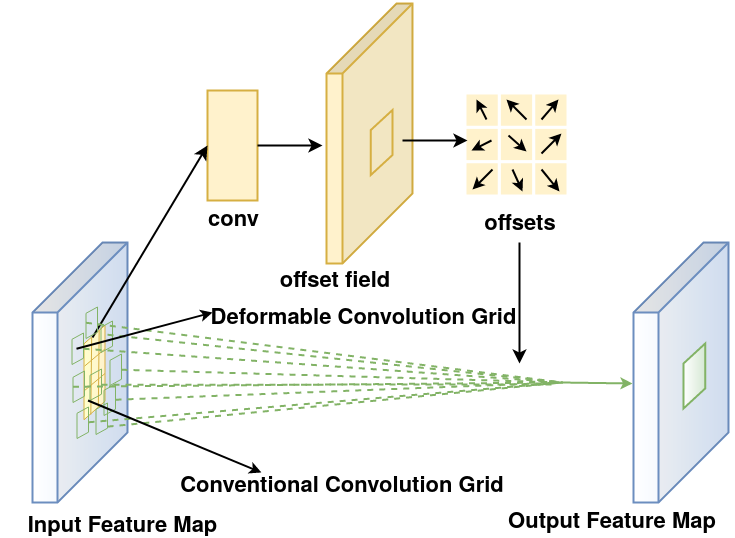, width=0.4\textwidth,height=0.27\textwidth}}}
\caption{Illustration of the deformable convolution. \label{fig:deformable}}
\end{figure}

 Deformable convolution is operated on R but with each points augmented by a learnable offset $\Delta p$. The offset value, $\Delta p$ is itself a trainable parameter. This helps in enabling each neuron to alter its receptive field based on the preceding feature map by creating an explicit offset. It makes the convolution operation agnostic for varying scales and transformations. The deformable convolution is shown in Figure~\ref{fig:deformable}.

\subsection{Implementation Details}

We implement {\sc cd}e{\sc c-n}et\footnote[2]{Our code is available publicly at \url{https://github.com/mdv3101/CDeCNet}} in Pytorch using {\sc mm}detection toolbox~\cite{mmdetection}. We use {\sc nvidia} {\sc g}e{\sc f}orce {\sc rtx} {\sc 2080 t}i {\sc gpu} with 12 {\sc gb} memory for our experiments. We use pre-trained {\sc r}es{\sc n}e{\sc x}t-101 (with blocks 3, 4, 23 and 3) on {\sc ms-coco}~\cite{lin2014microsoft} with {\sc fpn} as the network head. We train {\sc cd}e{\sc c-n}et with document images scaled to 1200 $\times$ 800, while maintaining the original aspect ratio, as the input. We use 0.00125 as an initial learning rate with a learning rate decay at 25 epoch and 40 epoch. We use 0.0033 as warmup schedule for first 500 iterations. {\sc cd}e{\sc c-n}et is trained for 50 epochs. However, for the larger datasets such as {\sc p}ub{\sc l}ay{\sc n}et and {\sc t}able{\sc b}ank, the model is trained for 8 epochs in total with learning rate decay at 4 epoch and 6 epoch. In case of fine-tuning, we use 12 epochs in total. We use three {\sc i}o{\sc u} threshold values --- 0.5, 0.6, and 0.7 in our model. We use 0.5, 1.0 and 2.0 as anchor ratio with a single anchor scale of 8. The batch size of 1 is used for training our models.    

\section{Experiments}

\subsection{Evaluation Measures}

Similar to the existing table localization tasks~\cite{gilani2017table,schreiber2017deepdesrt,siddiqui2018decnt,vo2018ensemble,arif2018table,li2019tablebank,sun2019faster,younas2019ffd,zhong2019publaynet,saha2019graphical,casado2019benefits} in document images, we also use recall, precision, F1, and mean average precision (m{\sc ap}) to evaluate the performance of {\sc cd}e{\sc c-n}et. For fair comparison, we evaluate the proposed {\sc cd}e{\sc c-n}et on same {\sc i}o{\sc u} threshold values as mentioned in the respective existing papers. We perform multi-scale testing at 7 different scales (with 3 smaller scales, original scale, and 3 larger scales). We select detection output as final result if it presents in at least 4 test cases out of 7 scales. It helps in eliminating the false positives and provide consistent results.

\subsection{Comparison with State-of-the-Arts on Benchmark Datasets}

\begin{table}[ht!]
\begin{center}
\begin{tabular}{|l|l|c c c c|} \hline
\textbf{Dataset} &\textbf{Method} &\multicolumn{4}{|c|}{\textbf{Score}} \\ \cline{3-6}
 &  &\textbf{R}$\uparrow$ &\textbf{P}$\uparrow$ &\textbf{F1}$\uparrow$ &\textbf{mAP}$\uparrow$ \\ \hline 
{\sc icdar-2013} &{\sc d}e{\sc cnt}~\cite{siddiqui2018decnt} &0.996 &0.996 &0.996 &- \\
 &{\sc cd}e{\sc c-n}et (our) &\textbf{1.000} &\textbf{1.000} &\textbf{1.000} &\textbf{1.000} \\ \hline
{\sc icadr-2017} &{\sc yolo}v3~\cite{huang2019yolo} &\textbf{0.968} &\textbf{0.975} &\textbf{0.971} &- \\
&{\sc cd}e{\sc c-n}et (our) &0.924 &0.970 &0.947 &\textbf{0.912} \\  \hline
{\sc icadr-2019} &{\sc t}able{\sc r}adar~\cite{gao2019icdar} &\textbf{0.940} &0.950 &\textbf{0.945} &- \\
 &{\sc cd}e{\sc c-n}et (our) &0.934 &\textbf{0.953} &0.944 &\textbf{0.922} \\ \hline 
{\sc unlv} &{\sc god}~\cite{saha2019graphical} &0.910 &0.946 &0.928 &- \\ 
 &{\sc cd}e{\sc c-n}et (our) &\textbf{0.925} &\textbf{0.952} &\textbf{0.938} & \textbf{0.912} \\ \hline
{\sc m}armot &{\sc d}e{\sc cnt}~\cite{siddiqui2018decnt} &\textbf{0.946} &0.849 &0.895 &- \\ 
&{\sc cd}e{\sc c-n}et (our) &0.930 &\textbf{0.975} &\textbf{0.952} &\textbf{0.911} \\ \hline
{\sc t}able{\sc b}ank &Li et al.~\cite{li2019tablebank} &0.975 &0.987 &0.981 &- \\
 &{\sc cd}e{\sc c-n}et (our) & \textbf{0.979} & \textbf{0.995} & \textbf{0.987} & \textbf{0.976}\\ \hline
{\sc p}ub{\sc l}ay{\sc n}et &{\sc m-rcnn}~\cite{zhong2019publaynet} &- & -&- &0.960 \\ 
 &{\sc cd}e{\sc c-n}et (our) & \textbf{0.970} & \textbf{0.988} & \textbf{0.978} & \textbf{0.967}  \\ \hline
 \end{tabular}
\end{center}
\caption{Illustrates comparison between {\sc cd}e{\sc c-n}et and state-of-the-art techniques on the existing benchmark datasets. \label{comparion_table}}
\end{table}

Comparison with current state-of-the-art techniques on various benchmark datasets is shown in Table~\ref{comparion_table}. We observe from the table that {\sc cd}e{\sc c-n}et outperforms state-of-the-art techniques on {\sc icadr-2013}, {\sc unlv}, {\sc m}armot, {\sc t}able{\sc b}ank, and {\sc p}ub{\sc l}ay{\sc n}et datasets. For {\sc icdar-2019}, {\sc cd}e{\sc c-n}et obtains very close performance to the state-of-the-art techniques. In case of {\sc icdar-2017} dataset, the performance of {\sc cd}e{\sc c-n}et is 2.4\% lower than the state-of-the-art method.

\begin{table*}
\addtolength{\tabcolsep}{-1.0pt}
\begin{center}
\begin{tabular}{|l| l | r|l |r|l|r| c| c c c c|} \hline
\textbf{Method} &\multicolumn{2}{|c|}{\textbf{Training}} &\multicolumn{2}{|c|}{\textbf{Fine-tuning}} &\multicolumn{2}{|c|}{\textbf{Test}} &\textbf{IoU} & \multicolumn{4}{|c|}{\textbf{Score}} \\ \cline{2-7} \cline{9-12}
  &\textbf{Dataset} &\textbf{\#Image} &\textbf{Dataset} &\textbf{\#Image} &\textbf{Dataset} &\textbf{\#Image} &  &\textbf{R}$\uparrow$ &\textbf{P}$\uparrow$ &\textbf{F1}$\uparrow$ &\textbf{mAP}$\uparrow$ \\ \hline
{\sc d}e{\sc cnt}~\cite{siddiqui2018decnt} &{\sc d}1 &4808 &- &- &{\sc icdar}-2013 &238 &0.5 &0.996$^{*}$ &0.996$^{*}$ &0.996$^{*}$ &- \\ 
{\sc cd}e{\sc c-n}et (our) &{\sc d}1 &4808 &- &- &{\sc icdar}-2013 &238 &0.5 &\textbf{1.000} &\textbf{1.000} &\textbf{1.000} &\textbf{1.000} \\ 
\hhline{|=|=|=|=|=|=|=|=|====|}
{\sc god}~\cite{saha2019graphical}   &Marmot &2K &- &- &{\sc icdar}-2013 &238 &0.5  &\textbf{1.000} &\textbf{0.982} &\textbf{0.991} &-  \\ 
{\sc cd}e{\sc c-n}et (our)         &Marmot &2K &-  &-  &{\sc icdar}-2013 &238 &0.5  &\textbf{1.000} &0.981 &\textbf{0.991} &\textbf{0.995}  \\ \hhline{|=|=|=|=|=|=|=|=|====|} 
{\sc f-rcnn}~\cite{zhong2019publaynet} &{\sc p}ub{\sc l}ay{\sc n}et &340K &{\sc icadr}-2013 &170 &{\sc icadr}-2013 &238 &0.5 &0.964 &0.972 &0.968 &  \\ 
{\sc m-rcnn}~\cite{zhong2019publaynet} &{\sc p}ub{\sc l}ay{\sc n}et &340K &{\sc icadr}-2013 &170 &{\sc icadr}-2013 &238 &0.5 &0.955 &0.940 &0.947 &-  \\ 
{\sc cd}e{\sc c-n}et (our) &{\sc p}ub{\sc l}ay{\sc n}et &340K &{\sc icadr}-2013 &170 &{\sc icadr}-2013 &238 &0.5 & \textbf{0.968}	& \textbf{0.987} & \textbf{0.977} & \textbf{0.959} \\ 
\hhline{|=|=|=|=|=|=|=|=|====|}  
{\sc yolo}v3+{\sc a}+{\sc pg}~\cite{huang2019yolo} &{\sc icdar}-2017 &1.6K &- &- &{\sc icadr}-2013 &238 &0.5 &0.949 &1.000 &0.973 &- \\ 
{\sc cd}e{\sc c-n}et (our)        &{\sc icdar}-2017 &1.6K &- &- &{\sc icadr}-2013 &238 &0.5 &\textbf{1.000} &\textbf{1.000} &\textbf{1.000} &\textbf{1.000} \\ \hhline{|=|=|=|=|=|=|=|=|====|} 
Khan et al.~\cite{khan2019table} &Marmot &2K &{\sc icdar}-2013 &204 &{\sc icdar}-2013 &34 &0.5 &0.901& 0.969& 0.934 &- \\   
{\sc t}able{\sc n}et+{\sc sf}~\cite{paliwal2019tablenet} &Marmot &2K &{\sc icdar}-2013 &204 &{\sc icdar}-2013 &34 &0.5 &0.963 & 0.970 & 0.966 &- \\ 
{\sc d}eep{\sc d}e{\sc srt}~\cite{schreiber2017deepdesrt} &Marmot &2K &{\sc icdar}-2013 &204 &{\sc icdar}-2013 &34 &0.5 &0.962 &0.974 &0.968 &- \\ 
{\sc cd}e{\sc c-n}et (our)                                                      &Marmot &2K &{\sc icdar}-2013 &204 &{\sc icdar}-2013 &34 &0.5 &\textbf{1.000} &\textbf{1.000} &\textbf{1.000} &\textbf{1.000} \\ 
\hhline{|=|=|=|=|=|=|=|=|====|} 
{\sc m-rcnn}~\cite{casado2019benefits} &Pascel {\sc voc} &16K &{\sc icdar}-2013 &178 &{\sc icdar}-2013 &60 &0.6 &0.770 &0.140 &0.230 &- \\
{\sc r}etina{\sc n}et~\cite{casado2019benefits} &Pascel {\sc voc} &16K &{\sc icdar}-2013 &178 &{\sc icdar}-2013 &60 &0.6 &0.580 &0.560 &0.570 &- \\ 
{\sc ssd}~\cite{casado2019benefits} &Pascel {\sc voc} &16K &{\sc icdar}-2013 &178 &{\sc icdar}-2013 &60 &0.6 &0.680 &0.540 &0.600 &- \\ 
{\sc yolo}~\cite{casado2019benefits} &Pascel {\sc voc} &16K &{\sc icdar}-2013 &178 &{\sc icdar}-2013 &60 &0.6 &0.580 &0.920 &0.750 &- \\     
{\sc cd}e{\sc c-n}et (our) &Pascel {\sc voc} &16K &{\sc icdar}-2013 &178 &{\sc icdar}-2013 &60 &0.6 &\textbf{0.844} &\textbf{1.000} &\textbf{0.922} &\textbf{0.844} \\ \hhline{|=|=|=|=|=|=|=|=|====|}
{\sc m-rcnn}~\cite{casado2019benefits} &{\sc t}able{\sc b}ank-{\sc l}a{\sc t}e{\sc x} &199K &{\sc icdar}-2013 &178 &{\sc icdar}-2013 &60 &0.6 &\textbf{0.970} &0.700 &0.810 &- \\   
{\sc r}etina{\sc n}et~\cite{casado2019benefits} &{\sc t}able{\sc b}ank-{\sc l}a{\sc t}e{\sc x} &199K &{\sc icdar}-2013 &178 &{\sc icdar}-2013 &60 &0.6 &0.770 &0.830 &0.800 &- \\   
{\sc ssd}~\cite{casado2019benefits} &{\sc t}able{\sc b}ank-{\sc l}a{\sc t}e{\sc x} &199K &{\sc icdar}-2013 &178 &{\sc icdar}-2013 &60 &0.6 &0.680 &0.620 &0.650 &- \\ 
{\sc yolo}~\cite{casado2019benefits}  &{\sc t}able{\sc b}ank-{\sc l}a{\sc t}e{\sc x} &199K &{\sc icdar}-2013 &178 &{\sc icdar}-2013 &60 &0.6 &0.650 &\textbf{1.000} &0.780 &- \\ 
{\sc cd}e{\sc c-n}et (our) &{\sc t}able{\sc b}ank-{\sc l}a{\sc t}e{\sc x} &199K &{\sc icdar}-2013 &178 &{\sc icdar}-2013 &60 &0.6 &0.933 &\textbf{1.000} &\textbf{0.967} &\textbf{0.933} \\ \hhline{|=|=|=|=|=|=|=|=|====|}
Kavasidis et al.~\cite{kavasidis2018saliencybased} &Custom dataset &45K &- &- &{\sc icdar}-2013 &238 &0.5 &0.981& 0.975& 0.978 &- \\ 
{\sc pftd}~\cite{melinda2019parameter} &- &- &- &- &{\sc icadr}-2013 &238 &0.5 &0.915& 0.939& 0.926& - \\  
Tran et al.~\cite{tran2015table} &- &- &- &- &{\sc icdar}-2013 &238 &0.5 &0.964 &0.952 &0.958 &- \\ 
 \hhline{|=|=|=|=|=|=|=|=|====|}
{\sc cd}e{\sc c-n}et$^{\ddagger}$ (our) &{\sc iiit-ar-13k} &9K &- &- &{\sc icdar}-2013 &238 &0.5 &0.942 &0.993 &0.968 &0.942 \\ \hline
\end{tabular}
\end{center}
\caption{Illustrates comparison between the proposed {\sc cd}e{\sc c-n}et and state-of-the-art techniques on {\sc icdar-2013} dataset. {\sc \textbf{a:}} indicates anchor optimization, {\sc \textbf{pg:}} indicates post-processing technique, {\sc \textbf{sf:}} indicates semantic features, {\sc \textbf{d1:}} indicates Marmot+{\sc unlv}+{\sc icdar-2017}, \textbf{*:} indicates the authors reported 0.996 in table however in discussion they mentioned 0.994. {\sc cd}\textbf{e}{\sc c-n}et$^{\ddagger}$\textbf{:} indicates a single  model which is trained with {\sc iiit-ar-13k} dataset. \label{table_icdar_2013}}
\end{table*}

\begin{table*}
\addtolength{\tabcolsep}{-3.1pt}
\begin{center}
\begin{tabular}{|l| l | r|l |r|l|r| c| c c c c|} \hline
\textbf{Method} &\multicolumn{2}{|c|}{\textbf{Training}} &\multicolumn{2}{|c|}{\textbf{Fine-tuning}} &\multicolumn{2}{|c|}{\textbf{Test}} &\textbf{IoU} & \multicolumn{4}{|c|}{\textbf{Score}} \\ \cline{2-7} \cline{9-12}
  &\textbf{Dataset} &\textbf{\#Image} &\textbf{Dataset} &\textbf{\#Image} &\textbf{Dataset} &\textbf{\#Image} &  &\textbf{R}$\uparrow$ &\textbf{P}$\uparrow$ &\textbf{F1}$\uparrow$ &\textbf{mAP}$\uparrow$ \\ \hline  
{\sc t}able{\sc r}adar~\cite{gao2019icdar} &{\sc icdar}-2019 &1200 &- &- &{\sc icdar}-2019 &439 &0.8 &\textbf{0.940} &0.950 &\textbf{0.945} &- \\
{\sc nlpr-pal}~\cite{gao2019icdar} &{\sc icdar}-2019 &1200 & & &{\sc icdar}-2019 &439 &0.8 &0.930 &0.930 &0.930 &- \\
{\sc l}enovo {\sc o}cean~\cite{gao2019icdar} &{\sc icdar}-2019 &1200 &- &- &{\sc icdar}-2019 &439 &0.8 &0.860 &0.880 &0.870 &- \\ 
{\sc cd}e{\sc c-n}et (our) &{\sc icdar}-2019 &1200 &- &- &{\sc icdar}-2019 &439 &0.8 &0.934 &\textbf{0.953} &0.944 &\textbf{0.922} \\ 
\hhline{|=|=|=|=|=|=|=|=|====|}
{\sc t}able{\sc r}adar~\cite{gao2019icdar} &{\sc icdar}-2019 &1200 &- &- &{\sc icdar}-2019 &439 &0.9 &0.890 &0.900 &0.895 &- \\
{\sc nlpr-pal}~\cite{gao2019icdar} &{\sc icdar}-2019 &1200 &- &- &{\sc icdar}-2019 &439 &0.9 &0.860 &0.860 &0.860 &- \\
{\sc l}enovo {\sc o}cean~\cite{gao2019icdar} &{\sc icdar}-2019 &1200 &- &- &{\sc icdar}-2019 &439 &0.9 &0.810 &0.820 &0.815 &- \\ 
{\sc cd}e{\sc c-n}et (our) &{\sc icdar}-2019 &1200 &- &- &{\sc icdar}-2019 &439 &0.9 &\textbf{0.904} &\textbf{0.922} &\textbf{0.913} &\textbf{0.843} \\ 
\hhline{|=|=|=|=|=|=|=|=|====|}
{\sc m-rcnn}~\cite{casado2019benefits} &Pascel {\sc voc} &16K &{\sc icdar}-2019 (archive) &599 &{\sc icdar}-2019 (archive) &198 &0.6 &0.640 &0.600 &0.620 &- \\
{\sc r}etina{\sc n}et~\cite{casado2019benefits} &Pascel {\sc voc} &16K &{\sc icdar}-2019 (archive) &599 &{\sc icdar}-2019 (archive) &198 &0.6 &0.660 &0.860 &0.740 &- \\
{\sc ssd}~\cite{casado2019benefits} &Pascel {\sc voc} &16K &{\sc icdar}-2019 (archive) &599 &{\sc icdar}-2019 (archive) &198 &0.6 &0.350 &0.310 &0.330 &- \\
{\sc yolo}~\cite{casado2019benefits} &Pascel {\sc voc} &16K &{\sc icdar}-2019 (archive) &599 &{\sc icdar}-2019 (archive) &198 &0.6 &0.910 &0.950 &0.930 &- \\ 
{\sc cd}e{\sc c-n}et (our) &Pascel {\sc voc} &16K &{\sc icdar}-2019 (archive) &599 &{\sc icdar}-2019 (archive) &198 &0.6 &\textbf{0.962} &\textbf{0.981} &\textbf{0.971} &\textbf{0.949} \\ 
\hhline{|=|=|=|=|=|=|=|=|====|} 
{\sc m-rcnn}~\cite{casado2019benefits} &{\sc t}able{\sc b}ank-{\sc l}a{\sc t}e{\sc x} &199K &{\sc icdar}-2019 (archive) &599 &{\sc icdar}-2019 (archive) &198 &0.6 &0.850 &0.760 &0.810 &- \\ 
{\sc r}etina{\sc n}et~\cite{casado2019benefits} &{\sc t}able{\sc b}ank-{\sc l}a{\sc t}e{\sc x} &199K &{\sc icdar}-2019 (archive) &599 &{\sc icdar}-2019 (archive) &198 &0.6 &0.740 &0.910 &0.820 &- \\ 
{\sc ssd}~\cite{casado2019benefits} &{\sc t}able{\sc b}ank-{\sc l}a{\sc t}e{\sc x} &199K &{\sc icdar}-2019 (archive) &599 &{\sc icdar}-2019 (archive) &198 &0.6 &0.350 &0.350 &0.350 &- \\ 
{\sc yolo}~\cite{casado2019benefits} &{\sc t}able{\sc b}ank-{\sc l}a{\sc t}e{\sc x} &199K &{\sc icdar}-2019 (archive) &599 &{\sc icdar}-2019 (archive) &198 &0.6 &0.950 &0.950 &0.950 &- \\  
{\sc cd}e{\sc c-n}et (our) &{\sc t}able{\sc b}ank-{\sc l}a{\sc t}e{\sc x} &199K &{\sc icdar}-2019 (archive) &599 &{\sc icdar}-2019 (archive) &198 &0.6 &\textbf{0.924} &\textbf{0.984} &\textbf{0.954} &\textbf{0.909} \\ \hhline{|=|=|=|=|=|=|=|=|====|}
{\sc cd}e{\sc c-n}et$^{\ddagger}$ (our) &{\sc iiit-ar-13k} &9K &- &- &{\sc icdar}-2019 &439 &0.8 & 0.625 &0.871 &0.748 &0.551 \\
{\sc cd}e{\sc c-n}et$^{\ddagger}$ (our) &{\sc iiit-ar-13k} &9K &{\sc icdar}-2019 &1200 &{\sc icdar}-2019 &439 &0.8 &0.930 &0.971 &0.950 &0.913 \\ \hline
\end{tabular}
\end{center}
\caption{Illustrates comparison between the proposed {\sc cd}e{\sc c-n}et and state-of-the-art techniques on {\sc icdar-2019} dataset. {\sc cd}e{\sc c-n}et$^{\ddagger}$\textbf{:} indicates a single  model which is trained with {\sc iiit-ar-13k} dataset. \label{table_icdar_2019}}
\end{table*}

\begin{table*}
\begin{center}
\begin{tabular}{|l| l | r|l |r|l|r| c| c c c c|} \hline
\textbf{Method} &\multicolumn{2}{|c|}{\textbf{Training}} &\multicolumn{2}{|c|}{\textbf{Fine-tuning}} &\multicolumn{2}{|c|}{\textbf{Test}} &\textbf{IoU} & \multicolumn{4}{|c|}{\textbf{Score}} \\ \cline{2-7} \cline{9-12}
  &\textbf{Dataset} &\textbf{\#Image} &\textbf{Dataset} &\textbf{\#Image} &\textbf{Dataset} &\textbf{\#Image} &  &\textbf{R}$\uparrow$ &\textbf{P}$\uparrow$ &\textbf{F1}$\uparrow$ &\textbf{mAP}$\uparrow$ \\ 
  \hline  
{\sc god}~\cite{saha2019graphical} &Marmot &2K &{\sc unlv} &340 &{\sc unlv} &84 &0.5 &0.910 &0.946 &0.928 &- \\ 
{\sc cd}e{\sc c-n}et (our) &Marmot &2K &{\sc unlv} &340 &{\sc unlv} &84 &0.5 &\textbf{0.925} &\textbf{0.952} &\textbf{0.938} & \textbf{0.912} \\ \hhline{|=|=|=|=|=|=|=|=|====|}
Gilani et al.~\cite{gilani2017table} &{\sc unlv} &340 &- &- &{\sc unlv} &84 &0.5 &\textbf{0.907} &0.823 &0.863 &- \\
{\sc cd}e{\sc c-n}et (our) &{\sc unlv} &340 &- &- &{\sc unlv} &84 &0.5 &0.906 &\textbf{0.914} &\textbf{0.910} &\textbf{0.861} \\ \hhline{|=|=|=|=|=|=|=|=|====|}
Arif and Shafait~\cite{arif2018table} &private &1019 &- &- &{\sc unlv} &427 &0.5 &\textbf{0.932} &0.863 &\textbf{0.896} &- \\ 
{\sc cd}e{\sc c-n}et (our) &private &1019 &- &- &{\sc unlv} &427 &0.5 &0.745 &\textbf{0.912} &0.829 &\textbf{0.711}  \\ \hhline{|=|=|=|=|=|=|=|=|====|}
{\sc d}e{\sc cnt}~\cite{siddiqui2018decnt} &{\sc d4} &4622 &- &- &{\sc unlv} &424 &0.5 &\textbf{0.749} &0.786 &0.767 &- \\ 
{\sc cd}e{\sc c-n}et (our) &{\sc d4} &4622 &- &- &{\sc unlv} &424 &0.5 &0.736 &\textbf{0.852} &\textbf{0.794} &\textbf{0.657} \\ \hhline{|=|=|=|=|=|=|=|=|====|}
{\sc m-rcnn}~\cite{casado2019benefits} &Pascel {\sc voc} &16K &{\sc unlv} &302 &{\sc unlv} &101 &0.6 &0.580 &0.290 &0.390 &- \\
{\sc r}etina{\sc n}et~\cite{casado2019benefits} &Pascel {\sc voc} &16K &{\sc unlv} &302 &{\sc unlv} &101 &0.6 &0.830 &0.810 &0.820 &- \\
{\sc ssd}~\cite{casado2019benefits} &Pascel {\sc voc} &16K &{\sc unlv} &302 &{\sc unlv} &101 &0.6 &0.640 &0.660 &0.650 &- \\
{\sc yolo}~\cite{casado2019benefits} &Pascel {\sc voc} &16K &{\sc unlv} &302 &{\sc unlv} &101 &0.6 &\textbf{0.950} &0.910 &\textbf{0.930} &- \\
{\sc cd}e{\sc c-n}et (our) &Pascel {\sc voc} &16K &{\sc unlv} &302 &{\sc unlv} &101 &0.6 &0.805 &\textbf{0.961} &0.883 &\textbf{0.788} \\ 
\hhline{|=|=|=|=|=|=|=|=|====|}
{\sc m-rcnn}~\cite{casado2019benefits} &{\sc t}able{\sc b}ank-{\sc l}a{\sc t}e{\sc x} &199K &{\sc unlv} &302 &{\sc unlv} &101 &0.6 &0.830 &0.660 &0.740 &- \\
{\sc r}etina{\sc n}et~\cite{casado2019benefits} &{\sc t}able{\sc b}ank-{\sc l}a{\sc t}e{\sc x} &199K &{\sc unlv} &302 &{\sc unlv} &101 &0.6 &0.830 &0.810 &0.820 &- \\
{\sc ssd}~\cite{casado2019benefits} &{\sc t}able{\sc b}ank-{\sc l}a{\sc t}e{\sc x} &199K &{\sc unlv} &302 &{\sc unlv} &101 &0.6 &0.660 &0.720 &0.690 &- \\
{\sc yolo}~\cite{casado2019benefits} &{\sc t}able{\sc b}ank-{\sc l}a{\sc t}e{\sc x} &199K &{\sc unlv} &302 &{\sc unlv} &101 &0.6 &\textbf{0.950} &0.930 &0.940 &- \\
{\sc cd}e{\sc c-n}et (our) &{\sc t}able{\sc b}ank-{\sc l}a{\sc t}e{\sc x} &199K &{\sc unlv} &302 &{\sc unlv} &101 &0.6 &0.894 &\textbf{0.991} &\textbf{0.943} &\textbf{0.889} \\
\hhline{|=|=|=|=|=|=|=|=|====|}  
{\sc cd}e{\sc c-n}et$^{\ddagger}$ (our) &{\sc iiit-ar-13k} &9K & &  &{\sc unlv} &424 &0.5 &0.770 &0.96 &0.865 &0.742 \\ 
{\sc cd}e{\sc c-n}et$^{\ddagger}$ (our) &{\sc iiit-ar-13k} &9K &private &1019  &{\sc unlv} &427 &0.5 & 0.776 & 0.958 & 0.866 & 0.750 \\ \hline
\end{tabular}
\end{center}
\caption{Illustrates comparison between the proposed {\sc cd}e{\sc c-n}et and state-of-the-art techniques on {\sc unlv} dataset. {\sc \textbf{d4:}} indicates {\sc icdar-2013}+{\sc icdar-2017}+Marmot. {\sc cd}e{\sc c-n}et$^{\ddagger}$\textbf{:} indicates a single  model which is trained with {\sc iiit-ar-13k} dataset. \label{table_unlv}}
\end{table*}

\begin{table*}[ht!]
\addtolength{\tabcolsep}{-2.0pt}
\begin{center}
\begin{tabular}{|l| l | r|l |r|l|r| c| c c c c|} \hline
\textbf{Method} &\multicolumn{2}{|c|}{\textbf{Training}} &\multicolumn{2}{|c|}{\textbf{Fine-tuning}} &\multicolumn{2}{|c|}{\textbf{Test}} &\textbf{IoU} & \multicolumn{4}{|c|}{\textbf{Score}} \\ \cline{2-7} \cline{9-12}
  &\textbf{Dataset} &\textbf{\#Image} &\textbf{Dataset} &\textbf{\#Image} &\textbf{Dataset} &\textbf{\#Image} &  &\textbf{R}$\uparrow$ &\textbf{P}$\uparrow$ &\textbf{F1}$\uparrow$ &\textbf{mAP}$\uparrow$ \\ \hline  
Li et al.~\cite{li2019tablebank} &{\sc t}able{\sc b}ank-{\sc l}a{\sc t}e{\sc x} &253K &- &- &{\sc t}able{\sc b}ank-{\sc w}ord &1K &0.5 & \textbf{0.956} &0.826 & \textbf{0.886} &- \\ 
 & & & & &{\sc t}able{\sc b}ank-{\sc l}a{\sc t}e{\sc x} &1K &0.5 &0.975 &0.987 &0.981 &- \\
 & & & & &{\sc t}able{\sc b}ank-both &2K &0.5 & \textbf{0.962} &0.872 &0.915 &- \\ \hline
{\sc cd}e{\sc c-n}et (our) &{\sc t}able{\sc b}ank-{\sc l}a{\sc t}e{\sc x} &253K &- &- &{\sc t}able{\sc b}ank-{\sc w}ord &1K &0.5 & 0.868 & \textbf{0.873} & 0.871 & \textbf{0.762} \\  
& & & & &{\sc t}able{\sc b}ank-{\sc l}a{\sc t}e{\sc x} &1K &0.5 & \textbf{0.979} & \textbf{0.995} & \textbf{0.987} & \textbf{0.976} \\
 & & & & &{\sc t}able{\sc b}ank-both &2K &0.5 & 0.924 & \textbf{0.934} & \textbf{0.929} & \textbf{0.898} \\ \hhline{|=|=|=|=|=|=|=|=|====|}
{\sc m-rcnn}~\cite{casado2019benefits} &{\sc t}able{\sc b}ank-{\sc l}a{\sc t}e{\sc x} &199K &- &- &{\sc t}able{\sc b}ank-{\sc l}a{\sc t}e{\sc x} &1K &0.6 &0.980 &0.960 &0.940 &- \\   
{\sc r}etina{\sc n}et~\cite{casado2019benefits} &{\sc t}able{\sc b}ank-{\sc l}a{\sc t}e{\sc x} &199K &- &- &{\sc t}able{\sc b}ank-{\sc l}a{\sc t}e{\sc x} &1K &0.6 &0.860 &0.980 &0.920 &- \\   
{\sc ssd}~\cite{casado2019benefits} &{\sc t}able{\sc b}ank-{\sc l}a{\sc t}e{\sc x} &199K &- &- &{\sc t}able{\sc b}ank-{\sc l}a{\sc t}e{\sc x} &1K &0.6 &0.970 &0.960 &0.965 &- \\ 
{\sc yolo}~\cite{casado2019benefits}  &{\sc t}able{\sc b}ank-{\sc l}a{\sc t}e{\sc x} &199K &- &- &{\sc t}able{\sc b}ank-{\sc l}a{\sc t}e{\sc x} &1K &0.6 & \textbf{0.990} &0.980 &0.985 &-\\ 
{\sc cd}e{\sc c-n}et (our)  &{\sc t}able{\sc b}ank-{\sc l}a{\sc t}e{\sc x} &199K &- &- &{\sc t}able{\sc b}ank-{\sc l}a{\sc t}e{\sc x} &1K &0.6 &0.978 &\textbf{0.995} &\textbf{0.986} &\textbf{0.974} \\ \hhline{|=|=|=|=|=|=|=|=|====|}
{\sc cd}e{\sc c-n}et$^{\ddagger}$ (our) &{\sc iiit-ar-13k} &9K & & &{\sc t}able{\sc b}ank-{\sc l}a{\sc t}e{\sc x} &1K &0.6 &0.779 &0.961 &0.870 &0.759 \\  
{\sc cd}e{\sc c-n}et$^{\ddagger}$ (our) &{\sc iiit-ar-13k} &9K &{\sc t}able{\sc b}ank-{\sc l}a{\sc t}e{\sc x} &199K &{\sc t}able{\sc b}ank-{\sc l}a{\sc t}e{\sc x} &1K &0.6 &0.970 &0.990 &0.980 &0.965 \\ \hline
\end{tabular}
\end{center}
\caption{Illustrates comparison between the proposed {\sc cd}e{\sc c-n}et (our) and state-of-the-art techniques on {\sc t}able{\sc b}ank dataset. {\sc cd}e{\sc c-n}et$^{\ddagger}$\textbf{:} indicates a single  model which is trained with {\sc iiit-ar-13k} dataset. \label{table_tablebank}}
\end{table*}

\subsection{Thorough Comparison with State-of-the-Art Techniques}

Tables~\ref{table_icdar_2013}-\ref{table_tablebank} presents the comparative results between the proposed {\sc cd}e{\sc c-n}et and the existing techniques on various benchmark datasets under the existing experimental environments. In most of the cases, {\sc cd}e{\sc c-n}et performs better than the existing techniques. The cascade Mask {\sc r-cnn} in {\sc cd}e{\sc c-n}et leads to significant reduction in number of false positives, which is evident from the high precision values. Table~\ref{table_icdar_2013} presents the obtained results under various experimental settings for {\sc icdar-2013}. We observe that for all experimental settings, {\sc cd}e{\sc c-n}et obtains the best results. In case of {\sc icdar-2019}, {\sc cd}e{\sc c-n}et performs only 0.1\% F1 score lower than state-of-the-art technique - {\sc t}able{\sc r}adar~\cite{gao2019icdar} at {\sc i}o{\sc u} threshold 0.8. At higher threshold value 0.9, {\sc cd}e{\sc c-n}et performs significantly (1.8\% greater F1 score) better than the state-of-the-art technique - {\sc t}able{\sc r}adar~\cite{gao2019icdar}. For all other experimental settings, {\sc cd}e{\sc c-n}et also obtain the best results. For {\sc unlv} dataset, {\sc cd}e{\sc c-n}et performs (2.7\% F1 score) better than the state-of-the-art method - {\sc d}e{\sc cnt}~\cite{siddiqui2018decnt}. For {\sc t}able{\sc b}ank dataset, {\sc cd}e{\sc c-n}et performs significantly  better that state-of-the-art technique - Li et al.~\cite{li2019tablebank}.              

\subsection{Effect of IoU Threshold on Table Detection}

We evaluate the trained {\sc cd}e{\sc c-n}et on the existing benchmark datasets under varying {\sc i}o{\sc u} thresholds to test robustness of the proposed network. Our experiments on various benchmark datasets shows that {\sc cd}e{\sc c-n}et gives consistent results over varying {\sc i}o{\sc u} thresholds. Table~\ref{table_iou_threshold} highlights that in case of {\sc icdar-2019} datasets, the {\sc cd}e{\sc c-n}et consistently obtains high detection accuracy under varying thresholds (in range 0.5-0.9). Our model also obtains consistent results (in range of 0.5-0.8) on {\sc icdar-2013} and {\sc unlv} datasets. Only at threshold 0.9, there is a performance drop on {\sc icdar-2013} and {\sc unlv} datasets. 


\begin{table}[ht!]
\addtolength{\tabcolsep}{-3.5pt}
\begin{center}
\begin{tabular}{|l|c c c|c c c|c c c|} \hline
\textbf{IoU} &\multicolumn{9}{c|}{\textbf{Performance on Various Benchmark Datasets}} \\ \cline{2-10}
\textbf{Threshold}  &\multicolumn{3}{c|}{\textbf{ICDAR-2013}} &\multicolumn{3}{c|}{\textbf{ICDAR-2019}}  &\multicolumn{3}{c|}{\textbf{UNLV}}  \\ \cline{2-10}   
 &\textbf{R}$\uparrow$ &\textbf{P}$\uparrow$ &\textbf{F1}$\uparrow$ &\textbf{R}$\uparrow$ &\textbf{P}$\uparrow$ &\textbf{F1}$\uparrow$ &\textbf{R}$\uparrow$ &\textbf{P}$\uparrow$ &\textbf{F1}$\uparrow$ \\ \hline
0.5 &1.000 &1.000 &1.000 &0.946 &0.987 &0.966 &0.770 &0.960 &0.865 \\
0.6 &1.000 &1.000 &1.000 &0.939 &0.980 &0.959 &0.758 &0.944 &0.851 \\ 
0.7 &0.987 &0.987 &0.987 &0.936 &0.977 &0.956 &0.734 &0.915 &0.825 \\
0.8 &0.942 &0.942 &0.942 &0.930 &0.971 &0.950 &0.663 &0.826 &0.744 \\
0.9 &0.660 &0.660 &0.660 &0.895 &0.934 &0.915 &0.496 &0.618 &0.557 \\ \hline
\end{tabular}
\end{center}
\caption{Illustrates the performance of {\sc cd}e{\sc c-n}et under varying {\sc i}o{\sc u} thresholds. \label{table_iou_threshold}}
\end{table}

\subsection{Qualitative Results}

\begin{figure*}[ht!]
\centerline{
\tcbox[sharp corners, size = tight, boxrule=0.2mm, colframe=black, colback=white]{
\psfig{figure=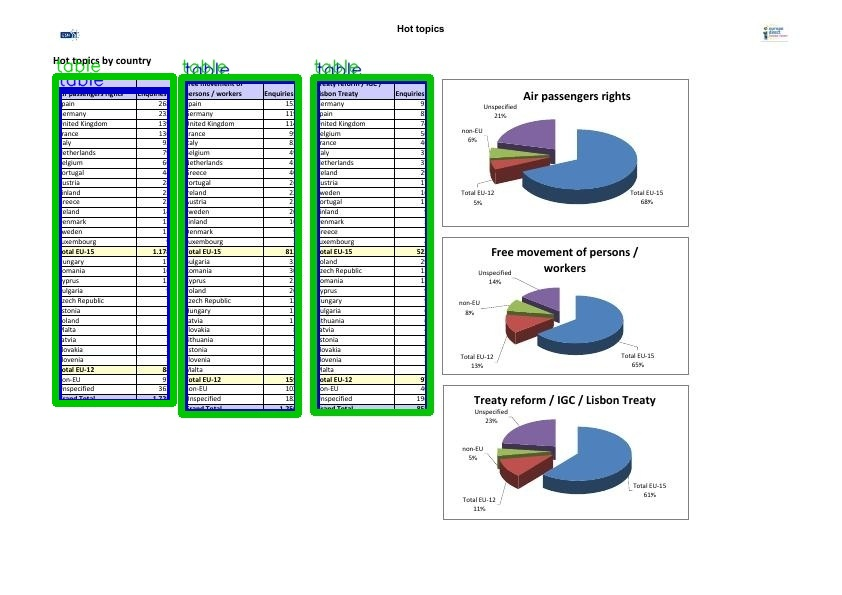, width=0.3\textwidth,height=0.3\textwidth}}
\hspace{-0.005\textwidth}
\tcbox[sharp corners, size = tight, boxrule=0.2mm, colframe=black, colback=white]{
\psfig{figure=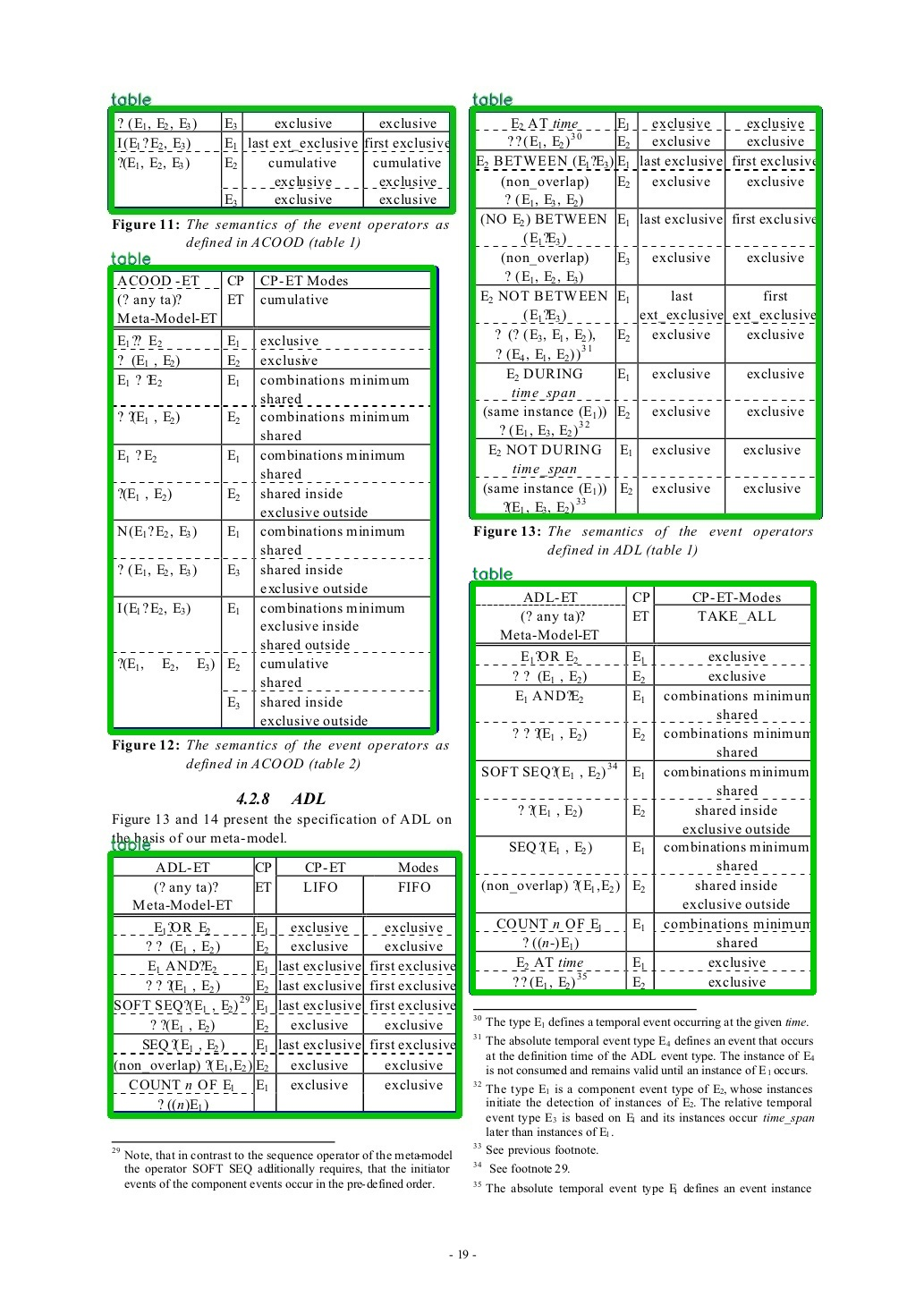, width=0.3\textwidth,height=0.3\textwidth}}
\hspace{-0.005\textwidth}
\tcbox[sharp corners, size = tight, boxrule=0.2mm, colframe=black, colback=white]{
\psfig{figure=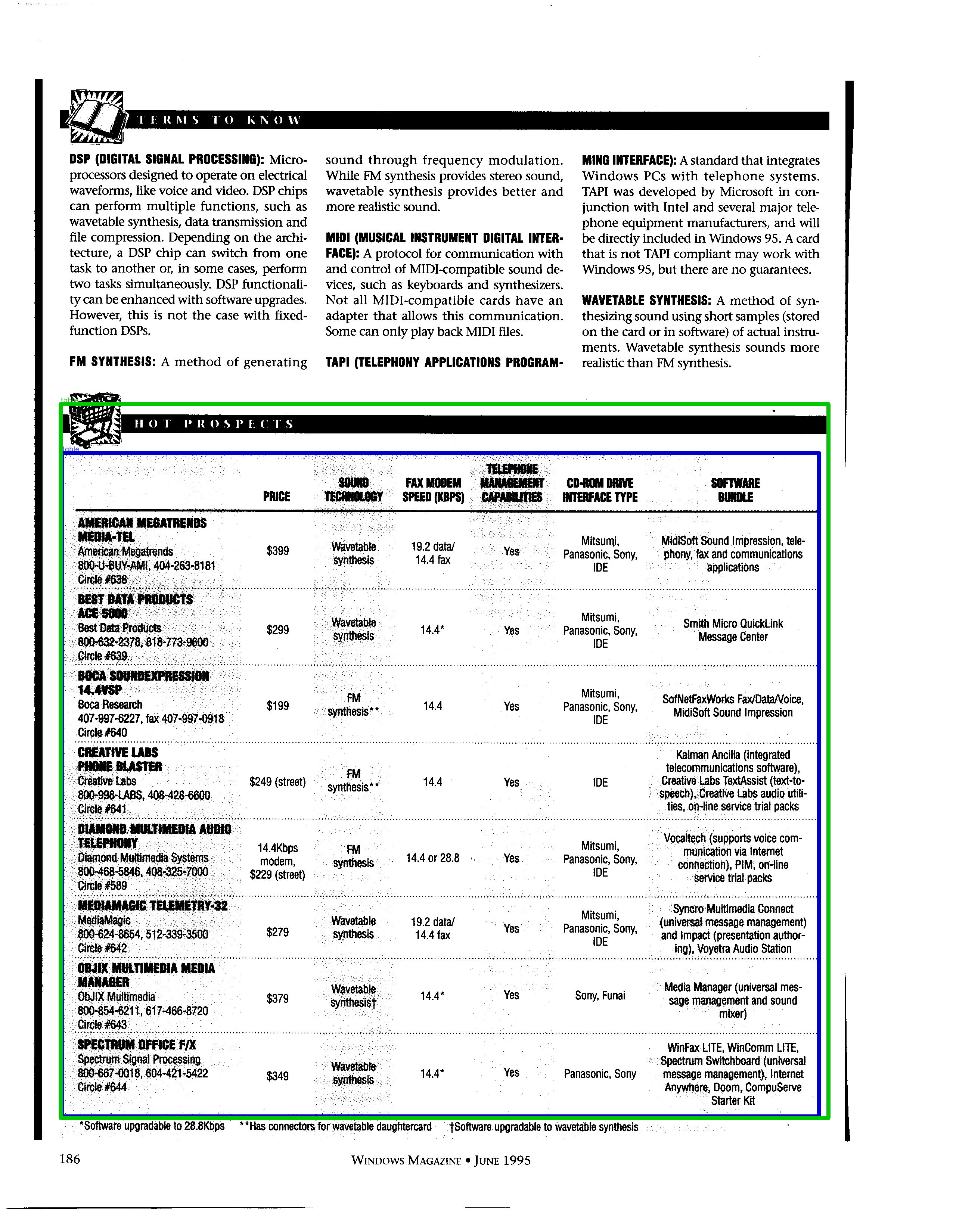, width=0.3\textwidth,height=0.3\textwidth}}}
\vspace{-0.01\textwidth}
\centerline{
\tcbox[sharp corners, size = tight, boxrule=0.2mm, colframe=black, colback=white]{
\psfig{figure=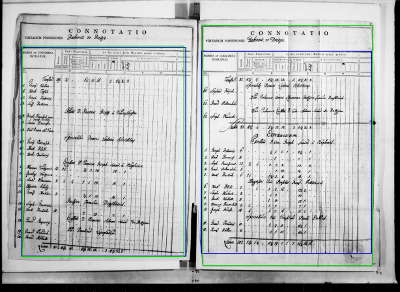, width=0.3\textwidth,height=0.3\textwidth}}
\hspace{-0.005\textwidth}
\tcbox[sharp corners, size = tight, boxrule=0.2mm, colframe=black, colback=white]{
\psfig{figure=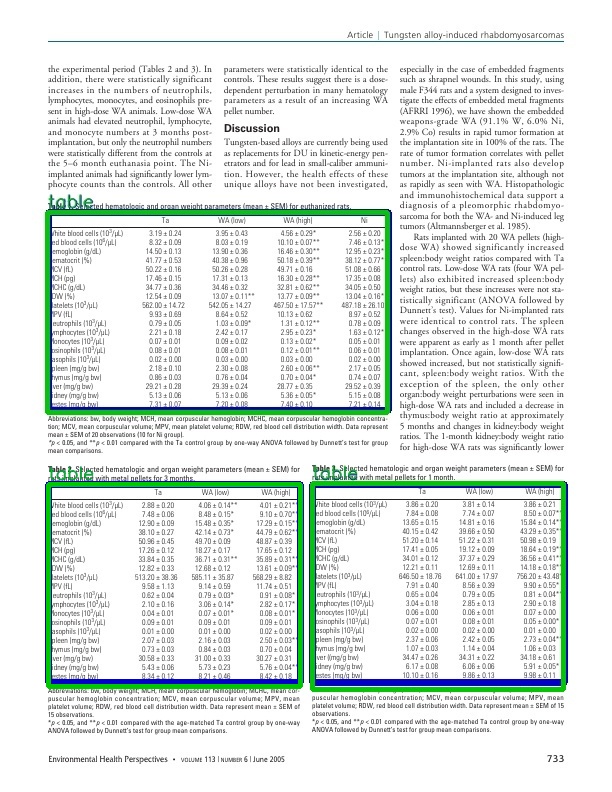, width=0.3\textwidth,height=0.3\textwidth}}
\hspace{-0.005\textwidth}
\tcbox[sharp corners, size = tight, boxrule=0.2mm, colframe=black, colback=white]{
\psfig{figure=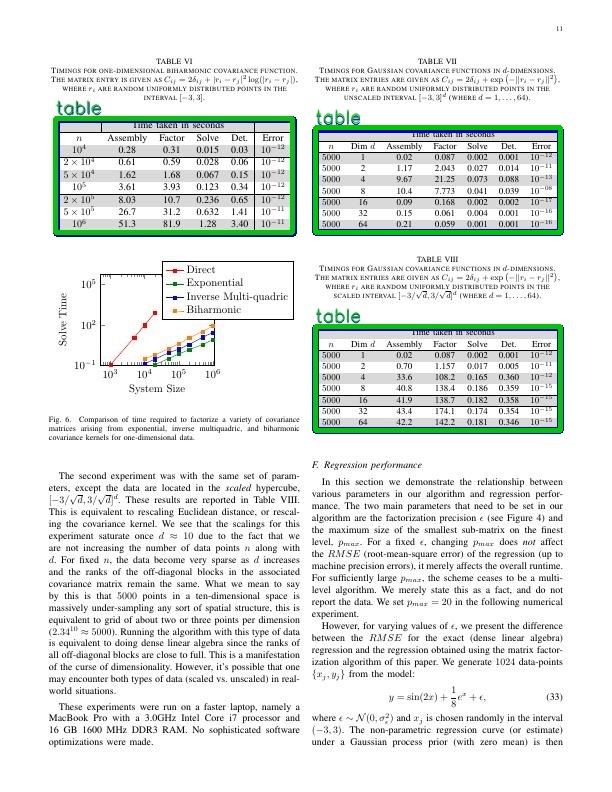, width=0.3\textwidth,height=0.3\textwidth}}}
\vspace{-0.01\textwidth}
\centerline{
\tcbox[sharp corners, size = tight, boxrule=0.2mm, colframe=black, colback=white]{
\psfig{figure=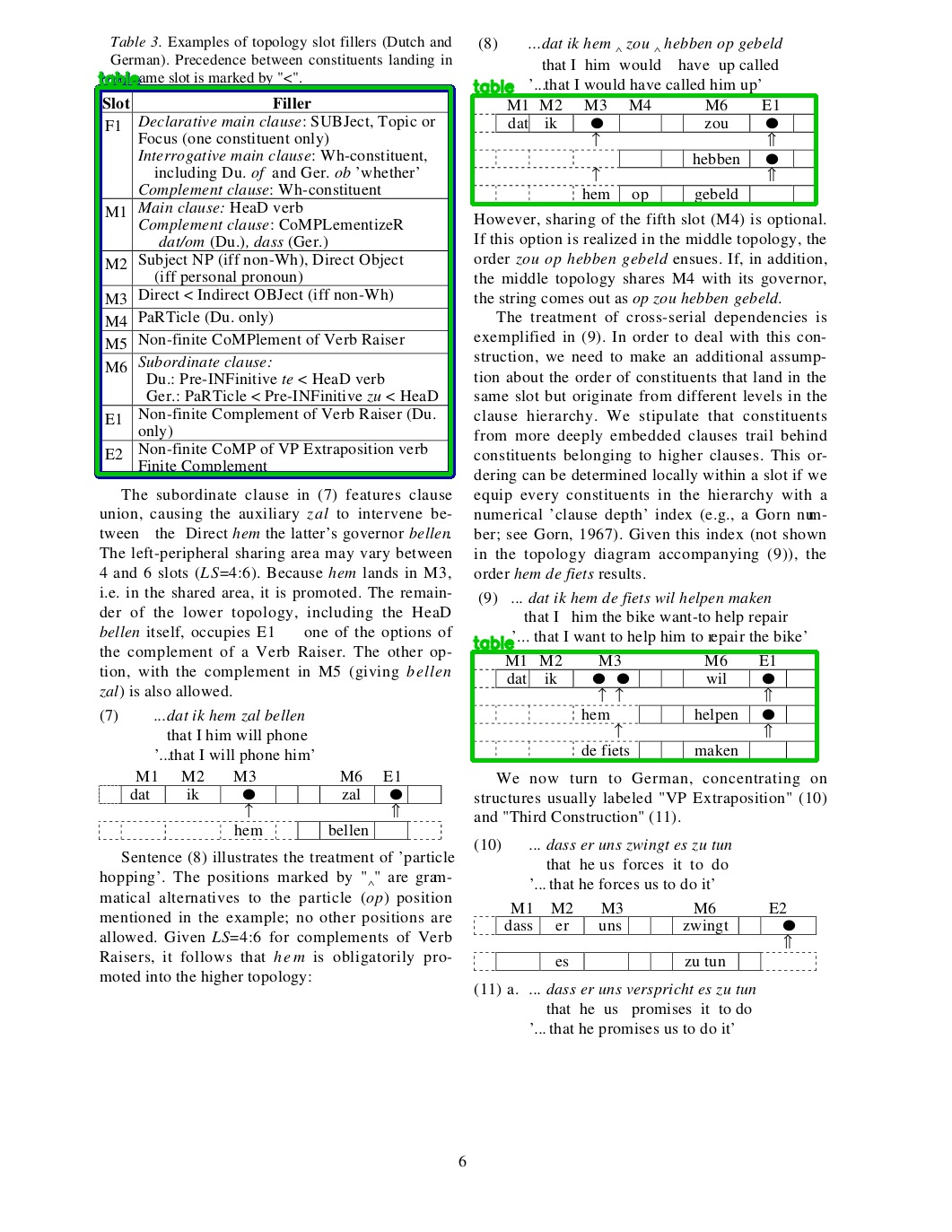, width=0.3\textwidth,height=0.3\textwidth}}
\hspace{-0.005\textwidth}
\tcbox[sharp corners, size = tight, boxrule=0.2mm, colframe=black, colback=white]{
\psfig{figure=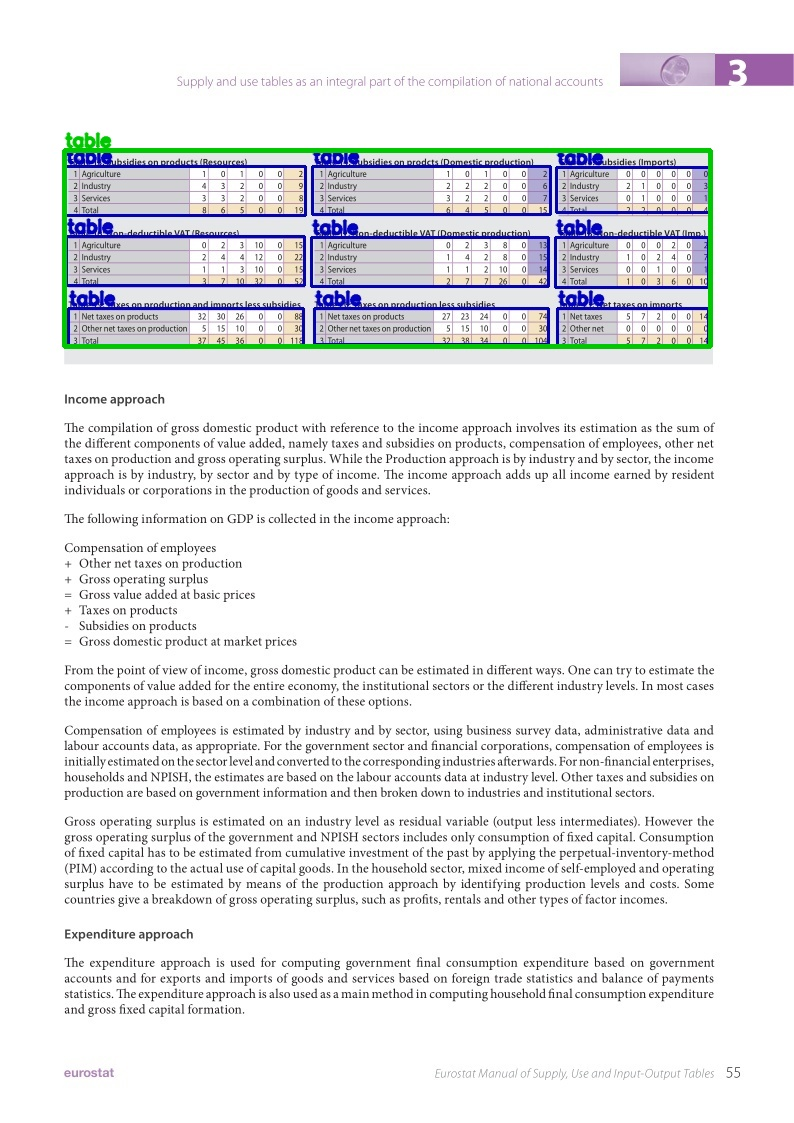, width=0.3\textwidth,height=0.3\textwidth}}
\hspace{-0.005\textwidth}
\tcbox[sharp corners, size = tight, boxrule=0.2mm, colframe=black, colback=white]{
\psfig{figure=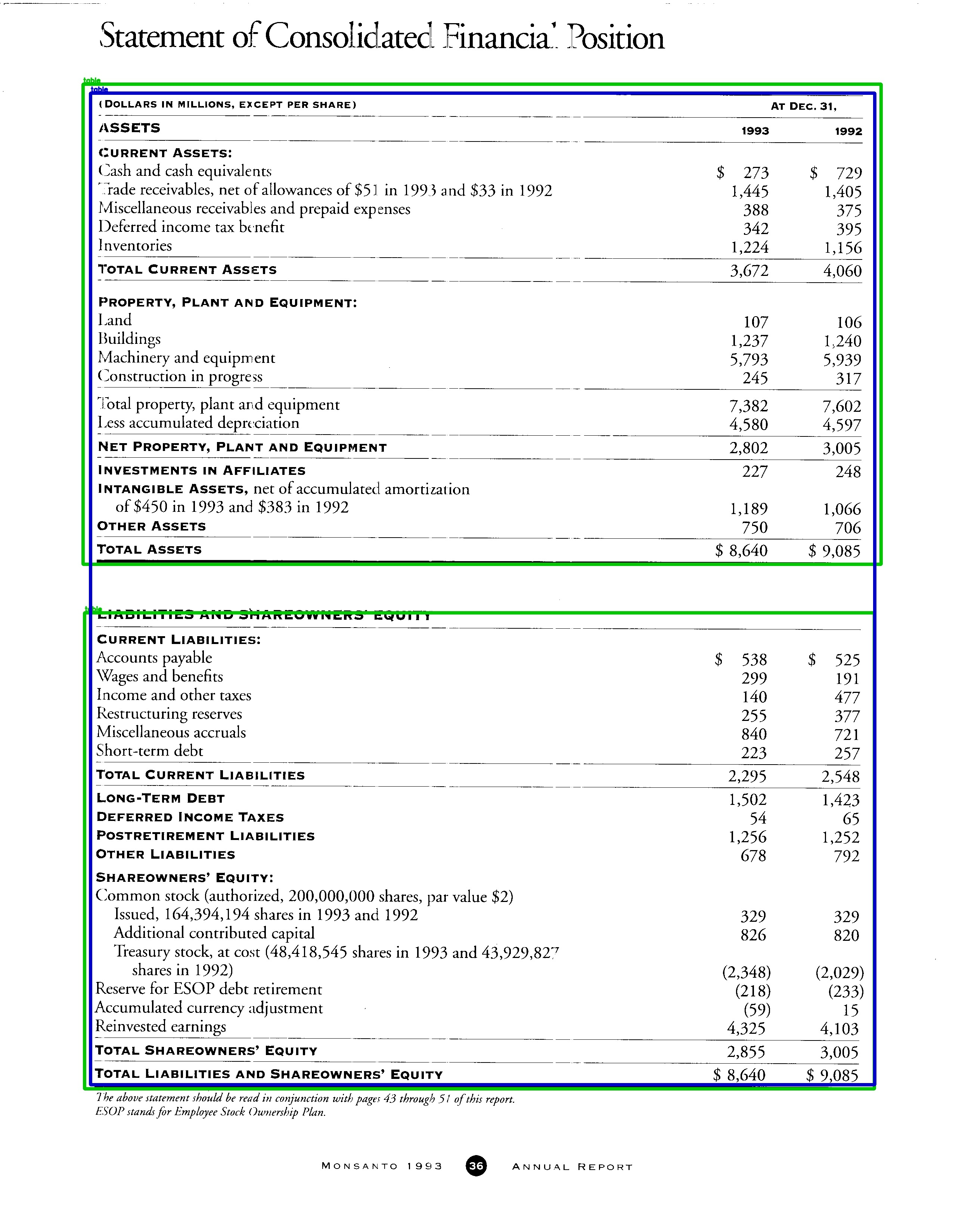, width=0.3\textwidth,height=0.3\textwidth}}}
\caption{Illustration of complex table detection results. Blue and Green colored rectangles correspond to ground truth and predicted bounding boxes using {\sc cd}e{\sc c-n}et. \textbf{First and Second Rows:} show examples where {\sc cd}e{\sc c-n}et accurately detects the tables. \textbf{Third Row:} shows examples where {\sc cd}e{\sc c-n}et fails to accurately detect the tables. \label{fig:quantitative_result}}
\end{figure*}

A visualization of detection results on {\sc icdar-2013}, {\sc icdar-pod-2017}, {\sc unlv} (first row, left to right),  {\sc icdar-2019} (c{\sc td}a{\sc r}), {\sc p}ub{\sc l}ay{\sc n}et and {\sc t}able{\sc b}ank (second row, left to right) obtained by {\sc cd}e{\sc c-n}et is shown in Figure~\ref{fig:quantitative_result}. The figure highlights that the {\sc cd}e{\sc c-n}et properly detects complex table with high confidence score.

Third row of Figure~\ref{fig:quantitative_result} shows some examples where {\sc cd}e{\sc c-n}et model fails to properly detect the tables. In the first image, it detects two false positives that are visually similar to tables. The second, and third images contain multiple closely spaced tables where {\sc cd}e{\sc c-n}et detects them as single table. 

\subsection{Results of Single Model} \label{result_single_model}

\begin{figure*}[ht!]
\centerline{
\tcbox[sharp corners, size = tight, boxrule=0.2mm, colframe=black, colback=white]{
\psfig{figure=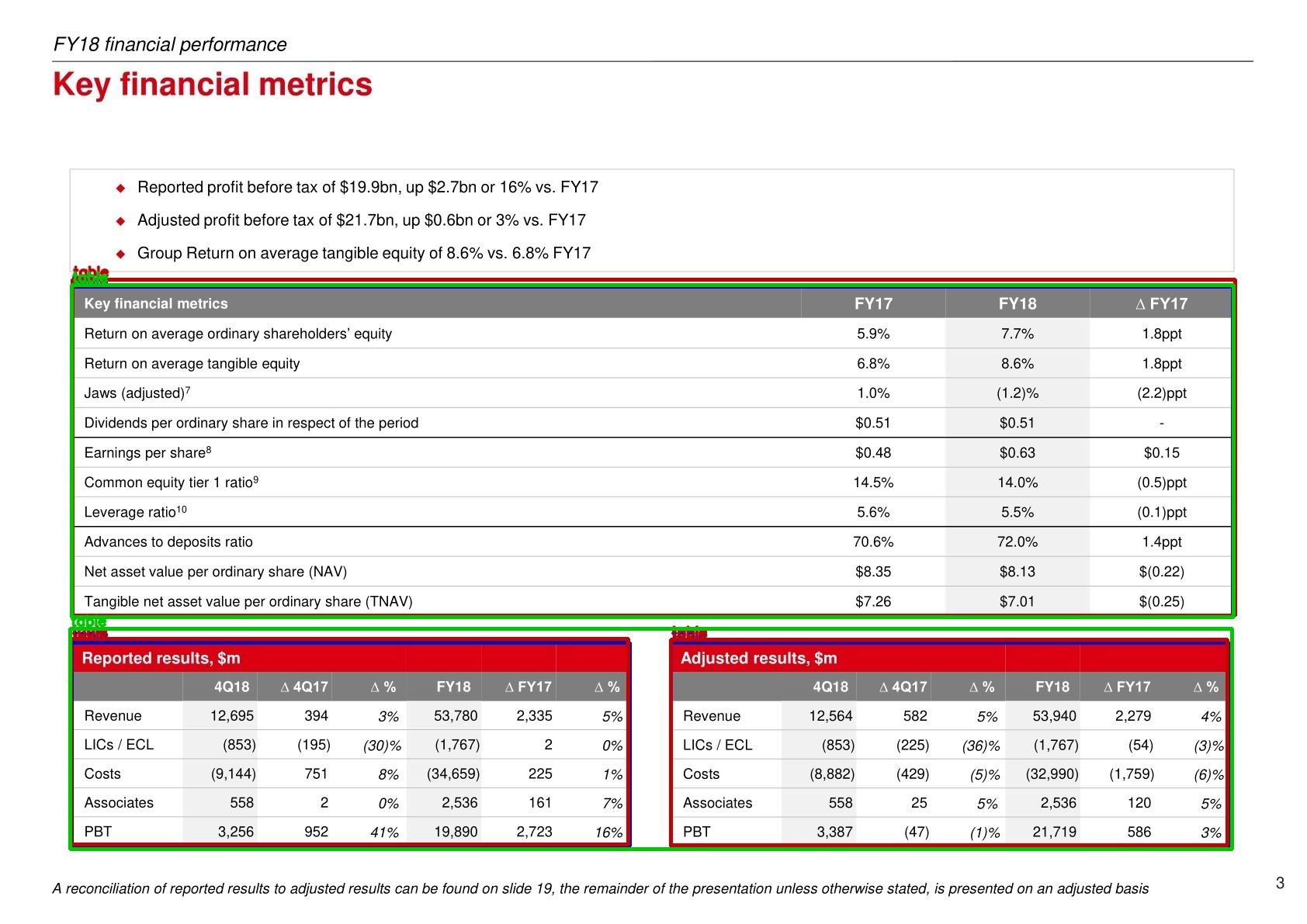, width=0.3\textwidth,height=0.3\textwidth}}
\hspace{-0.005\textwidth}
\tcbox[sharp corners, size = tight, boxrule=0.2mm, colframe=black, colback=white]{
\psfig{figure=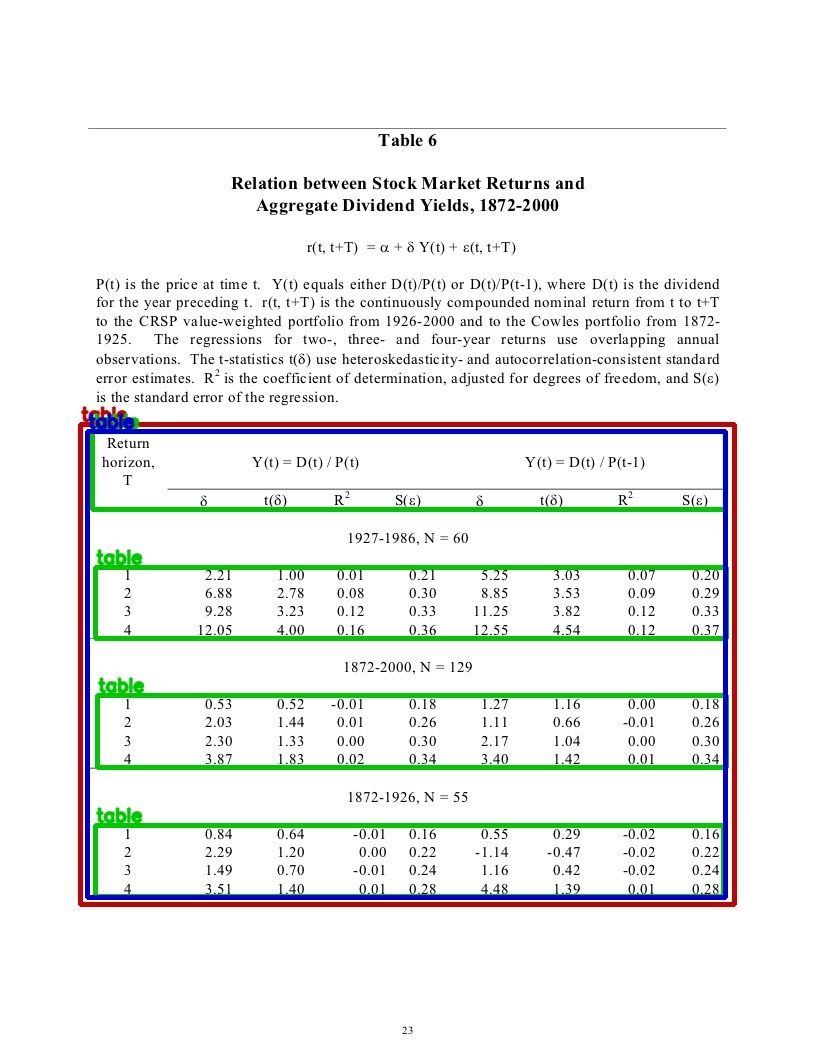, width=0.3\textwidth,height=0.3\textwidth}}
\hspace{-0.005\textwidth}
\tcbox[sharp corners, size = tight, boxrule=0.2mm, colframe=black, colback=white]{
\psfig{figure=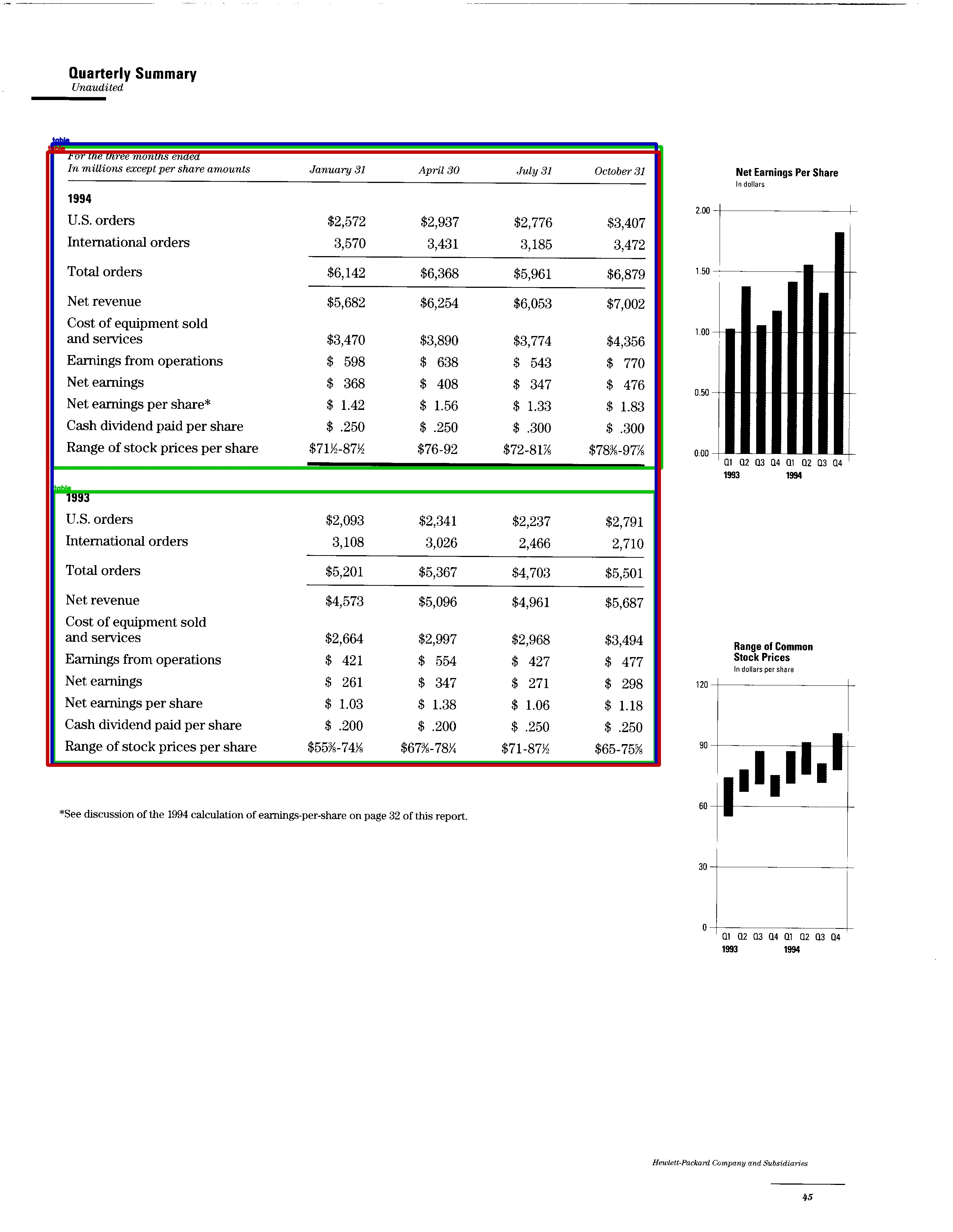, width=0.3\textwidth,height=0.3\textwidth}}}
\vspace{-0.01\textwidth}
\centerline{
\tcbox[sharp corners, size = tight, boxrule=0.2mm, colframe=black, colback=white]{
\psfig{figure=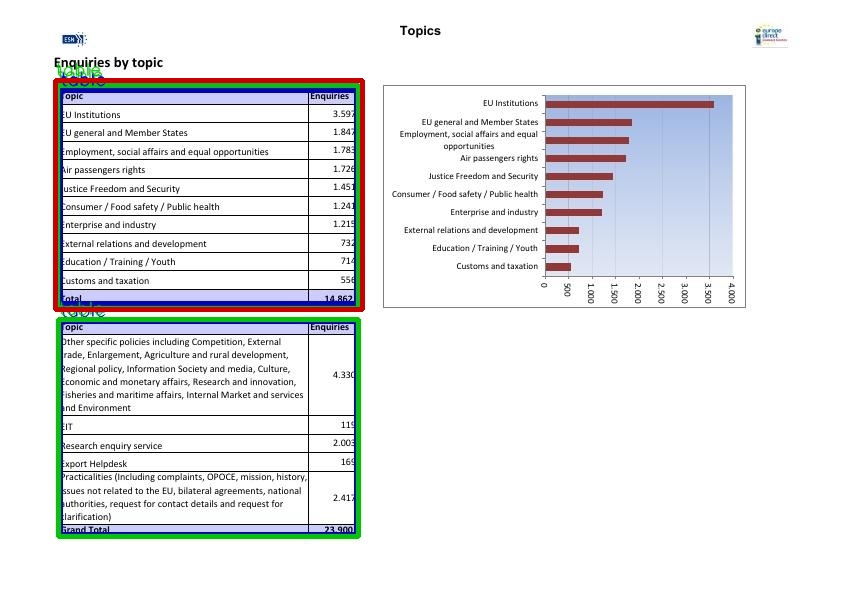, width=0.3\textwidth,height=0.3\textwidth}}
\hspace{-0.005\textwidth}
\tcbox[sharp corners, size = tight, boxrule=0.2mm, colframe=black, colback=white]{
\psfig{figure=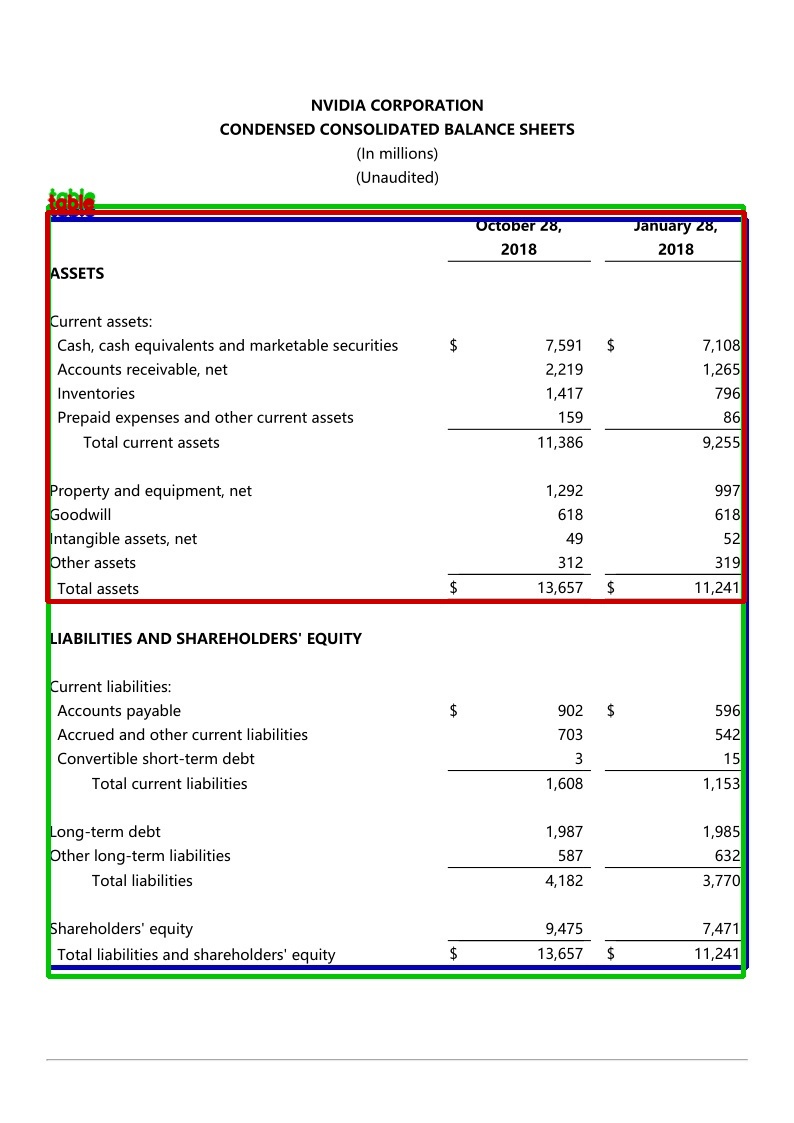, width=0.3\textwidth,height=0.3\textwidth}}
\hspace{-0.005\textwidth}
\tcbox[sharp corners, size = tight, boxrule=0.2mm, colframe=black, colback=white]{
\psfig{figure=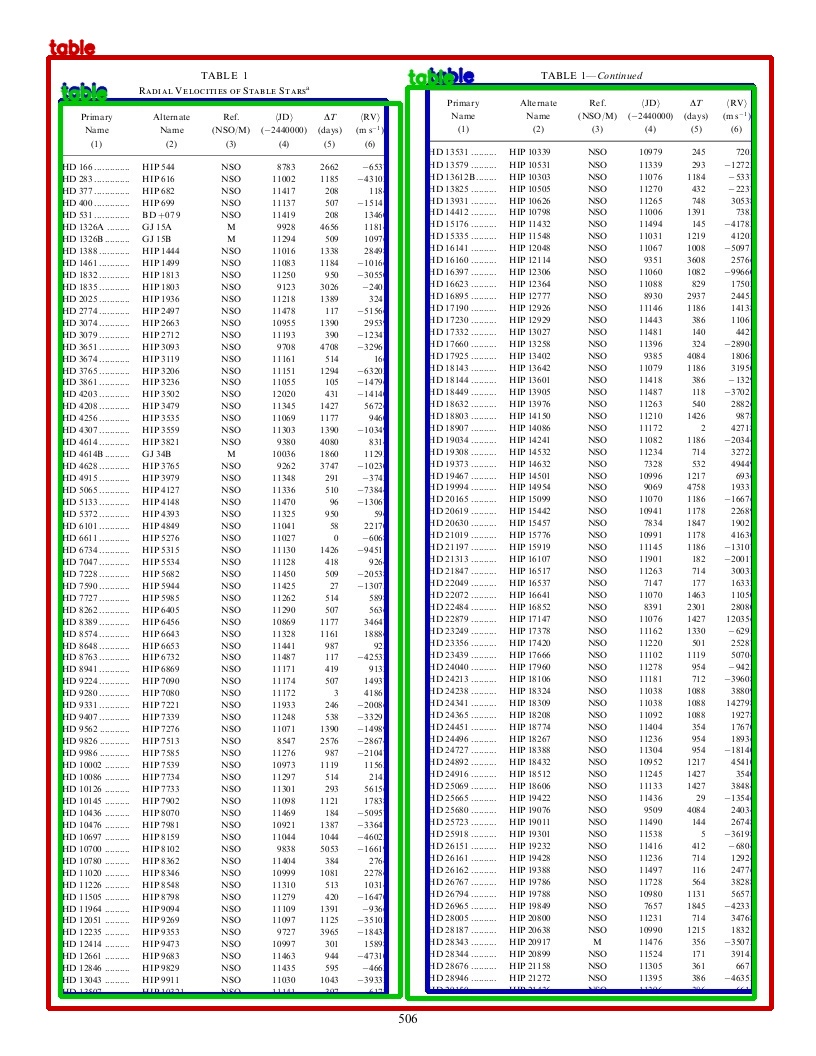, width=0.3\textwidth,height=0.3\textwidth}}}
\caption{Illustration visual results of the state-of-the-art {\sc cd}e{\sc c-n}et model and single {\sc cd}e{\sc c-n}et$^{\ddagger}$ model. Blue, Green, and Red colored rectangles correspond to ground truth and predicted bounding boxes using state-of-the-art {\sc cd}e{\sc c-n}et and single {\sc cd}e{\sc c-n}et$^{\ddagger}$ model respectively. \textbf{First Row:} shows examples where {\sc cd}e{\sc c-n}et$^{\ddagger}$ detects table accurately and {\sc cd}e{\sc c-n}et fails to detect table accurately. \textbf{Second Row:} shows examples where {\sc cd}e{\sc c-n}et detects table accurately and {\sc cd}e{\sc c-n}et$^{\ddagger}$ fails to detect table accurately. \label{fig:quantitative_single_result}}
\end{figure*}

Tables~\ref{table_icdar_2013}-\ref{table_tablebank} presents the comparative results between the proposed {\sc cd}e{\sc c-n}et and the existing techniques on various benchmark datasets under the existing experimental environments. The last row of each table presents obtained results using our single model {\sc cd}e{\sc c-n}et$^{\ddagger}$ trained with {\sc iiit-ar-13k} dataset, fine-tuned with training images and evaluated on test images of the respective datasets. Table~\ref{table_icdar_2013} highlights that our single model {\sc cd}e{\sc c-n}et$^{\ddagger}$ attains very close results to our best model {\sc cd}e{\sc c-n}et on {\sc icdar-2013} dataset. In case of {\sc icdar-2019}, our single model {\sc cd}e{\sc c-n}et$^{\ddagger}$ obtains the best performance at {\sc i}o{\sc u} threshold 0.8. In case of {\sc unlv} and {\sc t}able{\sc b}ank datasets, the performance of single model {\sc cd}e{\sc c-n}et$^{\ddagger}$ are very close to our best performing model {\sc cd}e{\sc c-n}et. 

Figure~\ref{fig:quantitative_single_result} presents the visual results obtained using our single model {\sc cd}e{\sc c-n}et$^{\ddagger}$ and our best model {\sc cd}e{\sc c-n}et. We select the best model under various existing experimental environments. First row of Figure~\ref{fig:quantitative_single_result} shows examples where single model {\sc cd}e{\sc c-n}et$^{\ddagger}$ performs better than the best model {\sc cd}e{\sc c-n}et. In those examples, our best model {\sc cd}e{\sc c-n}et predicts single bounding box for multiple tables. While single model {\sc cd}e{\sc c-n}et$^{\ddagger}$ accurately predicts bounding box corresponding to each table present in the document. The second row of Figure~\ref{fig:quantitative_single_result} presents examples where our best model {\sc cd}e{\sc c-n}et accurately detects all tables present in the documents. While our single model {\sc cd}e{\sc c-n}et$^{\ddagger}$ fails to predict bounding boxes corresponding to tables present in the documents. 

\subsection{Ablation Study}


\begin{table}[ht!]
\addtolength{\tabcolsep}{-4.0pt}
\begin{center}
\begin{tabular}{|l|c c c c|} \hline
\textbf{Models} &\multicolumn{4}{|c|}{\textbf{Score}} \\ \cline{2-5}
 &\textbf{R}$\uparrow$ &\textbf{P}$\uparrow$ &\textbf{F1}$\uparrow$ &\textbf{mAP}$\uparrow$ \\ \hline 
Cascade Mask {\sc r-cnn} with {\sc r}es{\sc n}e{\sc x}t-101 &0.987 &0.975 &0.981 &0.975 \\
 as backbone  &    &    &   & \\ \hline
Cascade Mask {\sc r-cnn} with composite &0.987 &0.981 &0.984 &0.973 \\
{\sc r}es{\sc n}e{\sc x}t-101 as backbone &   &  &   & \\ \hline
Cascade Mask {\sc r-cnn} with composite &\textbf{1.000} &\textbf{1.000} &\textbf{1.000} &\textbf{0.995}\\ 
{\sc r}es{\sc n}e{\sc x}t-101 having deformable convolution &  & &  & \\ 
as backbone (i.e., {\sc cd}e{\sc c-n}et) &  & &  & \\ \hline
\end{tabular}
\end{center}
\caption{Illustrates the performances of various models. All models are tested on {\sc icdar-2013} dataset with 0.5 as {\sc i}o{\sc u} threshold. Cascade Mask {\sc r-cnn} with composite {\sc r}es{\sc n}e{\sc x}t-101 having deformable convolution as backbone i.e., {\sc cd}e{\sc c-n}et obtains best results as compared to other models. We select {\sc cd}e{\sc c-n}et as our final model. \label{ablation_study_table}}
\end{table}
 
We perform a series of experiments to check the effectiveness of the proposed method. We train three models on {\sc m}armot dataset and evaluate on {\sc icdar}-2013. Our baseline model --- cascade Mask {\sc r-cnn} achieves F1 score of 0.981 at {\sc i}o{\sc u} threshold 0.5. We incorporate the dual backbone in the baseline model and obtain F1 score of 0.984. Again we incorporate deformable convolution instead of convolution in the dual backbone and call it as {\sc cd}e{\sc c-n}et, which attains the best F1 score of 1.000. This particular experiment highlights the utility of incorporating the key components --- dual backbone and deformable convolution into the baseline model Cascade Mask {\sc r-cnn}. We finally select {\sc cd}e{\sc c-n}et as our final model for table detection task.

\section{Conclusion} \label{conclusion}

We introduce a {\sc cd}e{\sc c}-{\sc n}et, which consists of a cascade Mask {\sc r-cnn} with a dual backbone having deformable convolution to detect tables present in documents with high accuracy at higher {\sc i}o{\sc u} threshold. The proposed {\sc cd}e{\sc c-n}et achieves state-of-the-art performance for most of the benchmark datasets under various existing experimental environments and significantly reduces the false positive detection even at the higher {\sc i}o{\sc u} threshold. We also provide a single model {\sc cd}e{\sc c}-{\sc n}et$^{\ddagger}$ for all benchmark datasets, which obtains very close performance to the state-of-the-art techniques. We expect that our single model sets a standard benchmark and improves the accuracy of table detection and other page objects-- figures, logos, mathematical expressions, etc. For future work, the current framework can be extended to a more challenging table structure recognition task. 

\bibliographystyle{IEEEtran}
\bibliography{reference}

\section*{Appendix A: Experiments}

\paragraph*{\textbf{Thorough Comparison with State-of-the-Art Techniques}}

We have done extensive experiments to check the performance of {\sc cd}e{\sc c-n}et. Our model is trained and evaluated on different experimental environments proposed by different researchers. The detailed results are shown in Tables~\ref{table_icdar_2017_arxiv}-\ref{table_marmot_arxiv}. To provide a fair comparison, we have used two strategies while training---(i) Train a model on the same dataset as proposed by a given researcher. The best model among them is later called as state-of-the-art model. (ii) Train a model on {\sc iiit-ar-13k} dataset and then evaluate the model directly on the dataset. After that, we fine-tune it using the training split of the dataset used by the respective researcher. The model is again evaluated on the testing split. This model is called as single {\sc cd}e{\sc c-n}et$^{\ddagger}$. The results are shown in the last rows of Tables~\ref{table_icdar_2017_arxiv}-\ref{table_marmot_arxiv}. It may be noted that for a given dataset, the single {\sc cd}e{\sc c-n}et$^{\ddagger}$ model which is trained on {\sc iiit-ar-13k} dataset and fine tuned only on the training split of the respective dataset (if available). 

{\sc cd}e{\sc c-n}et is trained and evaluated on {\sc icdar-2017} dataset (Table~\ref{table_icdar_2017_arxiv}). Our model does not achieve state-of-the-art performance. The main reason behind this is most of the methods used are quite focused on the {\sc icdar-2017} dataset, while {\sc cd}e{\sc c-n}et is generic in nature. We achieve the best result (F1 score of 0.959 at 0.6 {\sc i}o{\sc u} and 0.955 at 0.8 {\sc i}o{\sc u}) using the single {\sc cd}e{\sc c-n}et$^{\ddagger}$ model which is trained on {\sc iiit-ar-13k} and fine-tuned using {\sc icdar-2017} training dataset.

{\sc cd}e{\sc c-n}et was evaluated on {\sc m}armot dataset under various experimental environments and it achieves state-of-the-art results (Table~\ref{table_icdar_2017_arxiv}). The best performance was observed by the single {\sc cd}e{\sc c-n}et$^{\ddagger}$ model, which achieves an F1 score of 0.953.

Recently, two large datasets were released for table detection: {\sc t}able{\sc b}ank and {\sc p}ub{\sc l}ay{\sc n}et. We have evaluated the performance of {\sc cd}e{\sc c-n}et on them as well. {\sc cd}e{\sc c-n}et get an mAP of 0.967 when trained on {\sc p}ub{\sc l}ay{\sc n}et dataset and hence getting better results than the current benchmark as shown in table~\ref{table_publaynet_arxiv}. Our single model {\sc cd}e{\sc c-n}et$^{\ddagger}$ gets even better mAP score of 0.978.

\begin{table*}[ht!]
\addtolength{\tabcolsep}{-3.0pt}
\begin{center}
\begin{tabular}{|l| l | r|l |r|l|r| c| c c c c|} \hline
\textbf{Method} &\multicolumn{2}{|c|}{\textbf{Training}} &\multicolumn{2}{|c|}{\textbf{Fine-tuning}} &\multicolumn{2}{|c|}{\textbf{Test}} &\textbf{IoU} & \multicolumn{4}{|c|}{\textbf{Score}} \\ \cline{2-7} \cline{9-12}
  &\textbf{Dataset} &\textbf{\#Image} &\textbf{Dataset} &\textbf{\#Image} &\textbf{Dataset} &\textbf{\#Image} &  &\textbf{R}$\uparrow$ &\textbf{P}$\uparrow$ &\textbf{F1}$\uparrow$ &\textbf{mAP}$\uparrow$ \\ \hline
{\sc f}ast{\sc d}etectors$^{\dagger}$[16] &{\sc icdar-pod}-2017 &1600 &- &- &{\sc icdar-pod}-2017 &817 &0.6 &0.940 &0.903 &0.921 &0.925 \\
{\sc pal}$^{\dagger}$[16] &{\sc icdar-pod}-2017 &1600 &- &- &{\sc icdar-pod}-2017 &817 &0.6 &0.953 &0.968 &0.960 &0.933 \\ 
{\sc god}$^{\dagger}$[10] &{\sc icdar-pod}-2017 &1600 &- &- &{\sc icdar-pod}-2017 &817 &0.6 &- &- &0.971 &\textbf{0.989} \\ 
{\sc dsp-sc}$^{\dagger}$~\cite{li2018page} &{\sc icdar-pod}-2017 &1600 &- &- &{\sc icdar-pod}-2017 &817 &0.6 &0.962 &0.974 &0.968 &0.946 \\
{\sc yolo}v3$^{\dagger}$+{\sc a}+{\sc p}[18] &{\sc icdar-pod}-2017 &1600 &- &- &{\sc icdar-pod}-2017 &817 &0.6 &\textbf{0.972} &\textbf{0.978} &\textbf{0.975} &- \\ 
{\sc cd}e{\sc c-n}et$^{\dagger}$ (our) &{\sc icdar-pod}-2017 &1600 &- &- &{\sc icdar-pod}-2017 &817 &0.6 &0.931 & 0.977 & 0.954 & 0.920\\ \hhline{|=|=|=|=|=|=|=|=|====|}
{\sc f}ast{\sc d}etectors$^{\dagger}$[16] &{\sc icdar-pod}-2017 &1600 &- &- &{\sc icdar-pod}-2017 &817 &0.8 &0.915 &0.879 &0.896 &0.884 \\
{\sc pal}$^{\dagger}$[16] &{\sc icdar-pod}-2017 &1600 &- &- &{\sc icdar-pod}-2017 &817 &0.6 &0.943 &0.958 &0.951 &0.911 \\
{\sc god}$^{\dagger}$[10] &{\sc icdar-pod}-2017 &1600 &- & -&{\sc icdar-pod}-2017 &817 &0.8 &- &- &0.968 &\textbf{0.974} \\
{\sc dsp-sc}$^{\dagger}$~\cite{li2018page} &{\sc icdar-pod}-2017 &1600 &- &- &{\sc icdar-pod}-2017 &817 &0.8 &0.953 &0.965 &0.959 &0.923 \\ 
{\sc yolo}v3$^{\dagger}$+{\sc a}+{\sc p}[18] &{\sc icdar-pod}-2017 &1600 &- &- &{\sc icdar-pod}-2017 &817 &0.8 &\textbf{0.968} &\textbf{0.975} &\textbf{0.971} &- \\
{\sc cd}e{\sc c-n}et$^{\dagger}$ (our) &{\sc icdar-pod}-2017 &1600 &- &- &{\sc icdar-pod}-2017 &817 &0.8 &0.924 &0.970 &0.947 &0.912 \\ \hhline{|=|=|=|=|=|=|=|=|====|} 
{\sc d}e{\sc cnt}[3] &{\sc d}2 &4229 &- &- &{\sc icdar-pod}-2017 &817 &0.6 &\textbf{0.971} &0.965 &\textbf{0.968} &- \\ 
{\sc cd}e{\sc c-n}et (our) &{\sc d}2 &4229 &- &- &{\sc icdar-pod}-2017 &817 &0.6 &0.943 &\textbf{0.977} &0.960 &\textbf{0.938} \\ \hhline{|=|=|=|=|=|=|=|=|====|}
{\sc d}e{\sc cnt}[3] &{\sc d}2 &4229 &- &- &{\sc icdar-pod}-2017 &817 &0.8 &\textbf{0.937} &\textbf{0.967} &\textbf{0.952} &- \\ 
{\sc cd}e{\sc c-n}et (our) &{\sc d}2 &4229 &- &- &{\sc icdar-pod}-2017 &817 &0.8 &0.918 &0.951	&0.935 & \textbf{0.895} \\ 
 \hhline{|=|=|=|=|=|=|=|=|====|}
{\sc f}aster {\sc r-cnn}+{\sc cl}[4] &{\sc icdar-pod}-2017 &549 &- &- &{\sc icdar-pod}-2017 &243 &0.6 &\textbf{0.956} &0.943 &0.949 &- \\ 
{\sc f+m-rcnn}~\cite{li2019gan} &{\sc icdar-pod}-2017 &549 &- &- &{\sc icdar-pod}-2017 &243 &0.6 &0.944 &0.944 &0.944 &- \\ 
{\sc cd}e{\sc c-n}et (our) &{\sc icdar-pod}-2017 &549 &- &- &{\sc icdar-pod}-2017 &243 &0.6 &0.943 & \textbf{0.974} & \textbf{0.959} & \textbf{0.9308} \\ \hhline{|=|=|=|=|=|=|=|=|====|}
{\sc f+m-rcnn}~\cite{li2019gan} &{\sc icdar-pod}-2017 &549 &- &- &{\sc icdar-pod}-2017 &243 &0.8 &0.903 &0.903 &0.903 &- \\ 
{\sc cd}e{\sc c-n}et (our) &{\sc icdar-pod}-2017 &549 &- &- &{\sc icdar-pod}-2017 &243 &0.8 &\textbf{0.928} & \textbf{0.958} & \textbf{0.943} & \textbf{0.9023} \\ \hhline{|=|=|=|=|=|=|=|=|====|}
{\sc m-rcnn}[11] &Pascel {\sc voc} &16K &{\sc icdar-pod}-2017 &1200 &{\sc icdar-pod}-2017 &400 &0.6 &0.850 &0.320 &0.460 &- \\
{\sc r}etina{\sc n}et[11] &Pascel {\sc voc} &16K &{\sc icdar-pod}-2017 &1200 &{\sc icdar-pod}-2017 &400 &0.6 &0.860 &0.650 &0.740 &- \\ 
{\sc ssd}[11] &Pascel {\sc voc} &16K &{\sc icdar-pod}-2017 &1200 &{\sc icdar-pod}-2017 &400 &0.6 &0.710 &0.490 &0.580 &- \\ 
{\sc yolo}[11] &Pascel {\sc voc} &16K &{\sc icdar-pod}-2017 &1200 &{\sc icdar-pod}-2017 &400 &0.6 &\textbf{0.940} &0.900 &0.920 &- \\     
{\sc cd}e{\sc c-n}et (our) &Pascel {\sc voc} &16K &{\sc icdar-pod}-2017 &1200 &{\sc icdar-pod}-2017 &400 &0.6 & 0.932 &\textbf{0.981} &\textbf{0.956} &\textbf{0.925} \\ \hhline{|=|=|=|=|=|=|=|=|====|} 
{\sc m-rcnn}[11] &{\sc t}able{\sc b}ank-{\sc l}a{\sc t}e{\sc x} &199K &{\sc icdar-pod}-2017 &1200 &{\sc icdar-pod}-2017 &400 &0.6 &\textbf{0.950} &0.720 &0.820 &- \\ 
{\sc r}etina{\sc n}et[11]&{\sc t}able{\sc b}ank-{\sc l}a{\sc t}e{\sc x} &199K &{\sc icdar-pod}-2017 &1200 &{\sc icdar-pod}-2017 &400 &0.6 &0.870 &0.920 &0.890 &- \\ 
{\sc ssd}[11]&{\sc t}able{\sc b}ank-{\sc l}a{\sc t}e{\sc x} &199K &{\sc icdar-pod}-2017 &1200 &{\sc icdar-pod}-2017 &400 &0.6 &0.710 &0.550 &0.620 &- \\ 
{\sc yolo}[11] &{\sc t}able{\sc b}ank-{\sc l}a{\sc t}e{\sc x} &199K &{\sc icdar-pod}-2017 &1200 &{\sc icdar-pod}-2017 &400 &0.6 &0.940 &0.940 &0.940 &- \\ 
{\sc cd}e{\sc c-n}et (our) &{\sc t}able{\sc b}ank-{\sc l}a{\sc t}e{\sc x} &199K &{\sc icdar-pod}-2017 &1200 &{\sc icdar-pod}-2017 &400 &0.6 &0.914 &\textbf{0.980} &\textbf{0.947} &\textbf{0.905} \\ \hhline{|=|=|=|=|=|=|=|=|====|} 
{\sc cd}e{\sc c-n}et$^{\ddagger\dagger}$ (our) &{\sc iiit-ar-13k} &9K &-  &-  &{\sc icdar-pod}-2017 &817 &0.6 &0.776 &0.928 &0.852 &0.731 \\ 
{\sc cd}e{\sc c-n}et$^{\ddagger\dagger}$ (our) &{\sc iiit-ar-13k} &9K &{\sc icdar-pod}-2017 &1600 &{\sc icdar-pod}-2017 &817 &0.6 &0.931 &0.987 &0.959 &0.927 \\
{\sc cd}e{\sc c-n}et$^{\ddagger\dagger}$ (our) &{\sc iiit-ar-13k} &9K &-  &-  &{\sc icdar-pod}-2017 &817 &0.8 &0.625 &0.747 &0.686 &0.487 \\ 
{\sc cd}e{\sc c-n}et$^{\ddagger\dagger}$ (our) &{\sc iiit-ar-13k} &9K &{\sc icdar-pod}-2017 &1600 &{\sc icdar-pod}-2017 &817 &0.8 &0.928 &0.983 &0.955 &0.924 \\
{\sc cd}e{\sc c-n}et$^{\ddagger}$ (our) &{\sc iiit-ar-13k} &9K &{\sc d}2 &4229 &{\sc icdar-pod}-2017 &817 &0.6 &0.921 &0.957 &0.939 &0.897 \\ 
{\sc cd}e{\sc c-n}et$^{\ddagger}$ (our) &{\sc iiit-ar-13k} &9K &{\sc d}2 &4229 &{\sc icdar-pod}-2017 &817 &0.8 &0.909 &0.944 &0.926 &0.877 \\
{\sc cd}e{\sc c-n}et$^{\ddagger}$ (our) &{\sc iiit-ar-13k} &9K &- & -&{\sc icdar-pod}-2017 &243 &0.6 &0.751 &0.971 &0.861 &0.739 \\
{\sc cd}e{\sc c-n}et$^{\ddagger}$ (our) &{\sc iiit-ar-13k} &9K &{\sc icdar-pod}-2017 &549 &{\sc icdar-pod}-2017 &243 &0.6 &0.946 &0.984 &0.965 &0.934 \\
{\sc cd}e{\sc c-n}et$^{\ddagger}$ (our) &{\sc iiit-ar-13k} &9K &- & -&{\sc icdar-pod}-2017 &243 &0.8 &0.640 &0.829 &0.735 &0.549 \\
{\sc cd}e{\sc c-n}et$^{\ddagger}$ (our) &{\sc iiit-ar-13k} &9K &{\sc icdar-pod}-2017 &549 &{\sc icdar-pod}-2017 &243 &0.8 &0.937 &0.974 &0.955 &0.917 \\
{\sc cd}e{\sc c-n}et$^{\ddagger}$ (our) &{\sc iiit-ar-13k} &9K &- &- &{\sc icdar-pod}-2017 &400 &0.6 &0.772 &0.954 &0.863 &0.754 \\  
{\sc cd}e{\sc c-n}et$^{\ddagger}$ (our) &{\sc iiit-ar-13k} &9K &{\sc icdar-pod}-2017 &1200 &{\sc icdar-pod}-2017 &400 &0.6 &0.944 &0.975 &0.960 &0.930 \\ \hline
\end{tabular}
\end{center}
\caption{Illustrates comparison between the proposed {\sc cd}e{\sc c-n}et and state-of-the-art techniques on {\sc icdar-pod-2017}. {\sc \textbf{d2:}} indicates {\sc icdar-2013}+{\sc icdar-pod-2017}+{\sc unlv}+Marmot.\textbf{$\dagger$:} indicates model trained with multiple categories. {\sc cd}e{\sc c-n}et$^{\ddagger}$\textbf{:} indicates a single  model which is trained with {\sc iiit-ar-13k} dataset. \label{table_icdar_2017_arxiv}}
\end{table*}

\begin{table*}[ht!]
\begin{center}
\begin{tabular}{|l| l | r|l |r|l|r| c| c c c c|} \hline
\textbf{Method} &\multicolumn{2}{|c|}{\textbf{Training}} &\multicolumn{2}{|c|}{\textbf{Fine-tuning}} &\multicolumn{2}{|c|}{\textbf{Validation}} &\textbf{IoU} &\textbf{R}$\uparrow$ &\textbf{P}$\uparrow$ &\textbf{F1}$\uparrow$ &\textbf{mAP}$\uparrow$ \\ \cline{2-7}
  &\textbf{Dataset} &\textbf{\#Image} &\textbf{Dataset} &\textbf{\#Image} &\textbf{Dataset} &\textbf{\#Image} &  & & & & \\ \hline  
{\sc f-rcnn}$^{\dagger}$[9] &{\sc p}ub{\sc l}ay{\sc n}et &340K &- &- &{\sc p}ub{\sc l}ay{\sc n}et &11K &0.5-0.9 &- &- &- &0.954 \\ 
{\sc m-rcnn}$^{\dagger}$[9] &{\sc p}ub{\sc l}ay{\sc n}et &340K &- &- &{\sc p}ub{\sc l}ay{\sc n}et &11K &0.5-0.9  &- & -&- &0.960 \\ 
{\sc cd}e{\sc c-n}et$^{\dagger}$ (our) &{\sc p}ub{\sc l}ay{\sc n}et &340K &- &- &{\sc p}ub{\sc l}ay{\sc n}et &11K &0.5-0.9 & \textbf{0.970} & \textbf{0.988} & \textbf{0.978} & \textbf{0.967}
\\ \hhline{|=|=|=|=|=|=|=|=|====|}
{\sc cd}e{\sc c-n}et$^{\ddagger\dagger}$ (our) &{\sc iiit-ar-13k} &9K &- &- &{\sc p}ub{\sc l}ay{\sc n}et &11K &0.5-0.9  & 0.767 & 0.785 & 0.776 & 0.734 \\ 
{\sc cd}e{\sc c-n}et$^{\ddagger\dagger}$ (our) &{\sc iiit-ar-13k} &9K &{\sc p}ub{\sc l}ay{\sc n}et &340K &{\sc p}ub{\sc l}ay{\sc n}et &11K &0.5-0.9  & 0.975 & 0.993 & 0.984 & 0.978 \\ \hline
\end{tabular}
\end{center}
\caption{Illustrates comparison between the proposed {\sc cd}e{\sc c-n}et and state-of-the-art techniques on {\sc p}ub{\sc l}ay{\sc n}et dataset. \textbf{$\dagger$:} indicates model trained with multiple categories. {\sc cd}e{\sc c-n}et$^{\ddagger}$\textbf{:} indicates a single  model which is trained with {\sc iiit-ar-13k} dataset.\label{table_publaynet_arxiv}}
\end{table*}

\begin{table*}[ht!]
\addtolength{\tabcolsep}{-2.7pt}
\begin{center}
\begin{tabular}{|l| l | r|l |r|l|r| c| c c c c|} \hline
\textbf{Method} &\multicolumn{2}{|c|}{\textbf{Training}} &\multicolumn{2}{|c|}{\textbf{Fine-tuning}} &\multicolumn{2}{|c|}{\textbf{Test}} &\textbf{IoU} & \multicolumn{4}{|c|}{\textbf{Score}} \\ \cline{2-7} \cline{9-12}
  &\textbf{Dataset} &\textbf{\#Image} &\textbf{Dataset} &\textbf{\#Image} &\textbf{Dataset} &\textbf{\#Image} &  &\textbf{R}$\uparrow$ &\textbf{P}$\uparrow$ &\textbf{F1}$\uparrow$ &\textbf{mAP}$\uparrow$ \\ \hline  
{\sc d}e{\sc cnt}[3] &{\sc d3} &3079 &- &- &Marmot &1967 &0.5 &\textbf{0.946} &0.849 &0.895 &- \\ 
{\sc cd}e{\sc c-n}et (our) &{\sc d3} &3079 & & &Marmot &1967 &0.5 &0.930 &\textbf{0.975} &\textbf{0.952} &\textbf{0.911} \\ \hhline{|=|=|=|=|=|=|=|=|====|}
{\sc mfcn}+contour+{\sc crf}~\cite{he2017multi} &Various Doc &130 &- &- &Marmot &2000 &0.8 &0.731 &0.762 &0.747 &- \\ 
{\sc cd}e{\sc c-n}et (our) &Various Doc &130 & & &Marmot &2000 &0.8 &\textbf{0.836} &\textbf{0.845} &\textbf{0.840} &\textbf{0.716}  \\ \hhline{|=|=|=|=|=|=|=|=|====|}
{\sc mfcn}+contour+{\sc crf}~\cite{he2017multi} &Various Doc &130 &- &- &Marmot &2000 &0.9 &0.471 &0.481 &0.476 &- \\ 
{\sc cd}e{\sc c-n}et (our) &Various Doc &130 & & &Marmot &2000 &0.9 &\textbf{0.765} &\textbf{0.774} &\textbf{0.769} &\textbf{0.600} \\ \hhline{|=|=|=|=|=|=|=|=|====|}
{\sc m-rcnn}[11] &Pascel {\sc voc} &16K &Marmot (English) &744 &Marmot (English) &249 &0.6 &0.750 &0.370 &0.490 &- \\
{\sc r}etina{\sc n}et[11] &Pascel {\sc voc} &16K &Marmot (English) &744 &Marmot (English) &249 &0.6 &0.860 &0.750 &0.800 &- \\
{\sc ssd}[11] &Pascel {\sc voc} &16K &Marmot (English) &744 &Marmot (English) &249 &0.6 &0.760 &0.670 &0.710 &- \\
{\sc yolo}[11] &Pascel {\sc voc} &16K &Marmot (English) &744 &Marmot (English) &249 &0.6 &\textbf{0.960} &0.900 &0.930 &- \\
{\sc cd}e{\sc c-n}et (our) &Pascel {\sc voc} &16K &Marmot (English) &744 &Marmot (English) &249 &0.6 &0.946 &\textbf{0.993} &\textbf{0.969} &\textbf{0.942} \\ 
\hhline{|=|=|=|=|=|=|=|=|====|}
{\sc m-rcnn}[11] &{\sc t}able{\sc b}ank-{\sc l}a{\sc t}e{\sc x} &199K &Marmot (English) &744 &Marmot (English) &249 &0.6 &0.930 &0.720 &0.810 &- \\
{\sc r}etina{\sc n}et[11] &{\sc t}able{\sc b}ank-{\sc l}a{\sc t}e{\sc x} &199K &Marmot (English) &744 &Marmot (English) &249 &0.6 &0.860 &0.930 &0.900 &- \\
{\sc ssd}[11] &{\sc t}able{\sc b}ank-{\sc l}a{\sc t}e{\sc x} &199K &Marmot (English) &744 &Marmot (English) &249 &0.6 &0.750 &0.710 &0.730 &- \\
{\sc yolo}[11] &{\sc t}able{\sc b}ank-{\sc l}a{\sc t}e{\sc x} &199K &Marmot (English) &744 &Marmot (English) &249 &0.6 &\textbf{0.970} &0.950 &\textbf{0.960} &- \\
{\sc cd}e{\sc c-n}et (our) &{\sc t}able{\sc b}ank-{\sc l}a{\sc t}e{\sc x} &199K &Marmot (English) &744 &Marmot (English) &249 &0.6 &0.925 &\textbf{0.993} &0.959 &\textbf{0.924}  \\ \hhline{|=|=|=|=|=|=|=|=|====|}
{\sc m-rcnn}[11] &Pascel {\sc voc} &16K &Marmot (Chinese) &754 &Marmot (Chinese) &252 &0.6 &0.830 &0.520 &0.640 &- \\
{\sc r}etina{\sc n}et[11] &Pascel {\sc voc} &16K &Marmot (Chinese) &754 &Marmot (Chinese) &252 &0.6 &0.850 &0.780 &0.810 &- \\
{\sc ssd}[11] &Pascel {\sc voc} &16K &Marmot (Chinese) &754 &Marmot (Chinese) &252 &0.6 &0.700 &0.570 &0.630 &- \\
{\sc yolo}[11] &Pascel {\sc voc} &16K &Marmot (Chinese) &754 &Marmot (Chinese) &252 &0.6 &0.960 &0.950 &0.960 &- \\
{\sc cd}e{\sc c-n}et (our) &Pascel {\sc voc} &16K &Marmot (Chinese) &754 &Marmot (Chinese) &252 &0.6 &\textbf{0.966} &\textbf{0.988} &\textbf{0.977} &\textbf{0.959} \\ 
\hhline{|=|=|=|=|=|=|=|=|====|}
{\sc m-rcnn}[11] &{\sc t}able{\sc b}ank-{\sc l}a{\sc t}e{\sc x} &199K &Marmot (Chinese) &754 &Marmot (Chinese) &252 &0.6 &0.980 &0.820 &0.890 &- \\
{\sc r}etina{\sc n}et[11] &{\sc t}able{\sc b}ank-{\sc l}a{\sc t}e{\sc x} &199K &Marmot (Chinese) &754 &Marmot (Chinese) &252 &0.6 &0.870 &0.870 &0.870 &- \\
{\sc ssd}[11] &{\sc t}able{\sc b}ank-{\sc l}a{\sc t}e{\sc x} &199K &Marmot (Chinese) &754 &Marmot (Chinese) &252 &0.6 &0.670 &0.610 &0.640 &- \\
{\sc yolo}[11] &{\sc t}able{\sc b}ank-{\sc l}a{\sc t}e{\sc x} &199K &Marmot (Chinese) &754 &Marmot (Chinese) &252 &0.6 &0.930 &0.970 &0.950 &- \\
{\sc cd}e{\sc c-n}et (our) &{\sc t}able{\sc b}ank-{\sc l}a{\sc t}e{\sc x} &199K &Marmot (Chinese) &754 &Marmot (Chinese) &252 &0.6 &\textbf{0.966} &\textbf{0.994} &\textbf{0.980} &\textbf{0.962} \\ \hhline{|=|=|=|=|=|=|=|=|====|}
{\sc cd}e{\sc c-n}et$^{\ddagger}$ (our) &{\sc iiit-ar-13k} &9K &- &- &Marmot &1967 &0.5 &0.779 &0.943 &0.861 &0.756 \\
{\sc cd}e{\sc c-n}et$^{\ddagger}$ (our) &{\sc iiit-ar-13k} &9K &{\sc d3} &3079 &Marmot &1967 &0.5 &0.916 &0.991 &0.953 &0.909 \\ 
{\sc cd}e{\sc c-n}et$^{\ddagger}$ (our) &{\sc iiit-ar-13k} &9K &- &- &Marmot &2000 &0.8 &0.578 &0.682 &0.632 &0.427 \\ 
{\sc cd}e{\sc c-n}et$^{\ddagger}$ (our) &{\sc iiit-ar-13k} &9K &- &- &Marmot &2000 &0.9 &0.271 &0.322 &0.296 &0.108 \\ 
{\sc cd}e{\sc c-n}et$^{\ddagger}$ (our) &{\sc iiit-ar-13k} &9K &Various Doc &130 &Marmot &2000 &0.8 &0.833 &0.837 &0.835 &0.710 \\ 
{\sc cd}e{\sc c-n}et$^{\ddagger}$ (our) &{\sc iiit-ar-13k} &9K &Various Doc &130 &Marmot &2000 &0.9 &0.772 &0.775 &0.773 &0.603 \\
{\sc cd}e{\sc c-n}et$^{\ddagger}$ (our) &{\sc iiit-ar-13k} &9K &- &- &Marmot (English) &249 &0.6 &0.912 &0.964 &0.938 &0.906 \\
{\sc cd}e{\sc c-n}et$^{\ddagger}$ (our) &{\sc iiit-ar-13k} &9K &Marmot (English) &744 &Marmot (English) &249 &0.6 &0.952 &1.000 &0.976 &0.952 \\
{\sc cd}e{\sc c-n}et$^{\ddagger}$ (our) &{\sc iiit-ar-13k} &9K &- &- &Marmot (Chinese) &252 &0.6 &0.791 &0.921 &0.856 &0.736 \\
{\sc cd}e{\sc c-n}et$^{\ddagger}$ (our) &{\sc iiit-ar-13k} &9K &Marmot (Chinese) &754 &Marmot (Chinese) &252 &0.6 &0.944 &0.988 &0.966 &0.935 \\ \hline
\end{tabular}
\end{center}
\caption{Illustrates comparison between the proposed {\sc cd}e{\sc c-n}et and state-of-the-art techniques on {\sc m}armot dataset. {\sc \textbf{d3:}} indicates {\sc icdar-2013}+{\sc icdar-2017}+{\sc unlv}. {\sc cd}e{\sc c-n}et$^{\ddagger}$\textbf{:} indicates a single  model which is trained with {\sc iiit-ar-13k} dataset. \label{table_marmot_arxiv}}
\end{table*}

\paragraph*{\textbf{Effect of IoU Threshold on Table Detection}}

We evaluate the trained {\sc cd}e{\sc c-n}et on the existing benchmark datasets under varying {\sc i}o{\sc u} thresholds to test robustness of the proposed network. Our experiments on various benchmark datasets shows that {\sc cd}e{\sc c-n}et gives consistent results over varying {\sc i}o{\sc u} thresholds. Table~\ref{table_iou_threshold_appendix} highlights the obtained results under varying {\sc i}o{\sc u} thresholds using {\sc cd}e{\sc c-n}et.  

\begin{table*}[ht!]
\addtolength{\tabcolsep}{-3.5pt}
\begin{center}
\begin{tabular}{|l|c c c|c c c|c c c|c c c|c c c|c c c|c c c|} \hline
\textbf{IoU} &\multicolumn{21}{c|}{\textbf{Performance on Various Benchmark Datasets}} \\ \cline{2-22}
  &\multicolumn{3}{c|}{\textbf{ICDAR-2013}} &\multicolumn{3}{c|}{\textbf{ICDAR-2017}} &\multicolumn{3}{c|}{\textbf{ICDAR-2019}} &\multicolumn{3}{c|}{\textbf{Marmot}} &\multicolumn{3}{c|}{\textbf{UNLV}} &\multicolumn{3}{c|}{\textbf{TableBank}} &\multicolumn{3}{c|}{\textbf{PubLayNet}} \\ \cline{2-22}   
 &\textbf{R}$\uparrow$ &\textbf{P}$\uparrow$ &\textbf{F1}$\uparrow$ &\textbf{R}$\uparrow$ &\textbf{P}$\uparrow$ &\textbf{F1}$\uparrow$ &\textbf{R}$\uparrow$ &\textbf{P}$\uparrow$ &\textbf{F1}$\uparrow$ &\textbf{R}$\uparrow$ &\textbf{P}$\uparrow$ &\textbf{F1}$\uparrow$
 &\textbf{R}$\uparrow$ &\textbf{P}$\uparrow$ &\textbf{F1}$\uparrow$ &\textbf{R}$\uparrow$ &\textbf{P}$\uparrow$ &\textbf{F1}$\uparrow$ &\textbf{R}$\uparrow$ &\textbf{P}$\uparrow$ &\textbf{F1}$\uparrow$\\ \hline
0.5 &1.000 &1.000 &1.000 &0.934	&0.990 &0.962 &0.946 &0.987	&0.966 &0.916 &0.991 &0.953	&0.770 &0.960 &0.865 &0.979 &0.995 &0.987 &0.977 &0.996 &0.986 \\
0.6 &1.000 &1.000 &1.000 &0.931 &0.987 &0.959 &0.939 &0.980 &0.959 &0.911 &0.985 &0.948 &0.758 &0.944 &0.851 &0.977 &0.995 &0.986 &0.978 &0.995 &0.986 \\
0.7 &0.987 &0.987 &0.987 &0.931	&0.987 &0.959 &0.936 &0.977	&0.956 &0.905 &0.979 &0.942	&0.734 &0.915 &0.825 &0.978 &0.995 &0.986 &0.976 &0.994 &0.985\\
0.8 &0.942 &0.942 &0.942 &0.928	&0.983 &0.955 &0.930 &0.971	&0.950 &0.887 &0.960 &0.924	&0.663 &0.826 &0.744 &0.977 &0.993 &0.985 &0.974 &0.992 &0.983\\
0.9 &0.660 &0.660 &0.660 &0.902	&0.957 &0.929 &0.895 &0.934	&0.915 &0.823 &0.891 &0.857	&0.496 &0.618 &0.557 &0.966 &0.982 &0.974 &0.965 &0.983 &0.974\\ \hline
\end{tabular}
\end{center}
\caption{Illustrates the performance of {\sc cd}e{\sc c-n}et under varying {\sc i}o{\sc u} thresholds. \label{table_iou_threshold_appendix}}
\end{table*}

\end{document}